\documentclass[11pt]{article}
\pdfoutput=1
\usepackage{enumerate}
\usepackage{pdfsync}
\usepackage[OT1]{fontenc}

\usepackage[usenames]{color}
\usepackage[colorlinks,
            linkcolor=red,
            anchorcolor=blue,
            citecolor=blue
            ]{hyperref}
\usepackage{xcolor}   
\usepackage{colortbl}
\definecolor{LightCyan}{rgb}{0.8, 0.9, 1}
\usepackage{fullpage}
\usepackage[protrusion=true,expansion=true]{microtype}
\usepackage{pbox}
\usepackage{setspace}
\usepackage{tabularx}
\usepackage{float}
\usepackage{wrapfig,lipsum}
\usepackage{enumitem}
\usepackage{microtype}
\usepackage{graphicx}
\usepackage{subfigure}
\usepackage{enumerate}
\usepackage{enumitem}
\usepackage{pgfplots}
\usetikzlibrary{arrows,shapes,snakes,automata,backgrounds,petri}
\usepackage{booktabs} 

\usepackage{enumitem}

\usepackage{url}
\usepackage[utf8]{inputenc} 
\usepackage[T1]{fontenc}    
\usepackage{url}            
\usepackage{booktabs}       
\usepackage{amsfonts}       
\usepackage{amsmath}       
\usepackage{cleveref}
\usepackage{amsthm}
\newtheorem*{proposition*}{Proposition}
\usepackage{nicefrac}       
\usepackage[numbers]{natbib}
\usepackage{microtype}      
\usepackage{graphicx}
\usepackage[normalem]{ulem}
\usepackage{float}

\newcommand{\wba}{\mathbf{w}_{\alpha}}
\newcommand{\pdv}[2]{\frac{\partial #1}{ \partial #2}}
\newcommand{\val}{\text{val}}
\newcommand{\train}{\text{train}}
\newcommand{\test}{\text{test}}
\newcommand{\lin}{\text{lin}}

\renewcommand{\pdv}[2]{\frac{\partial #1}{ \partial #2}}

\renewcommand{\paragraph}[1]{\textbf{#1}}
\newcommand{\loo}{\text{\sc{loo}}}

\newcommand{\argmin}{\mathop{\mathrm{argmin}}}

\newcommand{\pp}[1]{\textbf{Proof of Proposition #1:\\}}

\RequirePackage{amsmath}
\RequirePackage{amssymb}
\RequirePackage{amsthm}
\RequirePackage{bm} 
\RequirePackage{url}
\usepackage{multirow}
\usepackage{natbib}
\usepackage{graphicx}
\usepackage{subfigure}
\usepackage{makecell}
\usepackage{booktabs}
\usepackage{array}
\usepackage{algorithm}
\usepackage{algorithmic}
\usepackage{dsfont}

\let\hat\widehat

\newcommand{\fb}{\mathbf{f}}

\newcommand{\wb}{\mathbf{w}}
\newcommand{\xb}{\mathbf{x}}
\newcommand{\yb}{\mathbf{y}}
\newcommand{\zb}{\mathbf{z}}

\newcommand{\Ab}{\mathbf{A}}
\newcommand{\Bb}{\mathbf{B}}
\newcommand{\Cb}{\mathbf{C}}
\newcommand{\Db}{\mathbf{D}}

\newcommand{\Ib}{\mathbf{I}}

\newcommand{\Kb}{\mathbf{K}}
\newcommand{\Lb}{\mathbf{L}}

\newcommand{\Rb}{\mathbf{R}}

\newcommand{\Xb}{\mathbf{X}}
\newcommand{\Yb}{\mathbf{Y}}
\newcommand{\Zb}{\mathbf{Z}}

\newcommand{\cD}{\mathcal{D}}
\newcommand{\cE}{\mathcal{E}}

\newcommand{\cT}{{\mathcal{T}}}

\newcommand{\RR}{\mathbb{R}}

\newcommand{\diag}{{\rm diag}}

\newtheoremstyle{mytheoremstyle} 
    {\topsep}                    
    {\topsep}                    
    {\normalfont}                   
    {}                           
    {\bfseries}                   
    {.}                          
    {.5em}                       
    {}  

\theoremstyle{mytheoremstyle}

\ifx\BlackBox\undefined
\newcommand{\BlackBox}{\rule{1.5ex}{1.5ex}}  
\fi

\ifx\QED\undefined
\def\QED{~\rule[-1pt]{5pt}{5pt}\par\medskip}
\fi

\ifx\proof\undefined
\newenvironment{proof}{\par\noindent{\bf Proof\ }}{\hfill\BlackBox\\[2mm]}
\fi

\ifx\theorem\undefined
\newtheorem{theorem}{Theorem}
\fi
\ifx\example\undefined
\newtheorem{example}[theorem]{Example}
\fi
\ifx\property\undefined
\newtheorem{property}{Property}
\fi
\ifx\lemma\undefined
\newtheorem{lemma}[theorem]{Lemma}
\fi
\ifx\proposition\undefined
\newtheorem{proposition}[theorem]{Proposition}
\fi
\ifx\remark\undefined
\newtheorem{remark}[theorem]{Remark}
\fi
\ifx\corollary\undefined
\newtheorem{corollary}[theorem]{Corollary}
\fi
\ifx\definition\undefined
\newtheorem{definition}[theorem]{Definition}
\fi
\ifx\conjecture\undefined
\newtheorem{conjecture}[theorem]{Conjecture}
\fi
\ifx\fact\undefined
\newtheorem{fact}[theorem]{Fact}
\fi
\ifx\claim\undefined
\newtheorem{claim}[theorem]{Claim}
\fi
\ifx\assumption\undefined
\newtheorem{assumption}[theorem]{Assumption}
\fi

\allowdisplaybreaks

\title{\huge DIVA: Dataset Derivative of a Learning Task}

\def\affiliation#1{\gdef\@affiliation{#1}} \gdef\@affiliation{}

\begin{document}

\author{Yonatan Dukler$^{1,2}$\thanks{Work conducted at Amazon Web Services.}\ , Alessandro Achille$^1$, Giovanni Paolini$^1$, Avinash Ravichandran$^1$, \\ Marzia Polito$^1$, Stefano Soatto$^1$\\
$^1$ Amazon Web Services, \\
\texttt{\{aachille, paoling, ravinash, mpolito, soattos\}@amazon.com}\\
$^2$ 
Department of Mathematics, \\ 
University of California, Los Angeles\\
\texttt{ydukler@math.ucla.edu}\\
}
\date{}
\maketitle

\begin{abstract}
We present a method to compute the derivative of a learning task with respect to a dataset. A learning task is a function from a training set to the validation error, which can be represented by a trained deep neural network (DNN). The ``dataset derivative'' is a linear operator, computed around the trained model, that informs how perturbations of the weight of each training sample affect the validation error, usually computed on a separate validation dataset.  Our method, DIVA (Differentiable Validation) hinges on a closed-form differentiable expression of the leave-one-out cross-validation error around a pre-trained DNN. Such expression constitutes the dataset derivative. DIVA could be used for dataset auto-curation, for example removing samples with faulty annotations, augmenting a dataset with additional relevant samples, or rebalancing. More generally, DIVA can be used to optimize the dataset, along with the parameters of the model, as part of the training process without the need for a separate validation dataset, unlike bi-level optimization methods customary in AutoML. To illustrate the flexibility of DIVA, we report experiments on sample auto-curation tasks such as outlier rejection, dataset extension, and automatic aggregation of multi-modal data.
\end{abstract}

\bibliographystyle{plainnat}

\section{Introduction}
\label{sec:introduction}
Consider the following seemingly disparate questions. 
{\em (i) Dataset Extension:}  Given a relatively small training set, but access to a large pool of additional data, how to select from the latter samples to augment the former? 
{\em (ii) Dataset Curation:} Given a potentially large dataset riddled with annotation errors,
how to automatically reject such outlier samples? {\em (iii) Dataset Reweighting:} 
Given a finite training set, how to reweight the training samples to yield better generalization performance?

These three are examples of {\em Dataset Optimization}. In order to solve this problem with differentiable programming, one can optimize a loss of the model end-to-end, which requires {\em differentiating the model's loss with respect to the dataset.} Our main contribution is an efficient method to compute such a {\em dataset derivative}. This allows learning an importance weight $\alpha_i$ for each datum in a training dataset $\cD$, extending the optimization from the weights $\wb$ of a parametric model such as a deep neural network (DNN), to also include the weights of the dataset.

As illustrated in the following diagram, standard optimization in machine learning works by finding the weights $\wb_{\alpha}$ that minimize the training loss $L_\text{train}(\wb, D_\alpha) = \sum_i \alpha_i \ell(f_\wb(\xb_i), y_i)$ on a given (weighted) dataset $D_\alpha$ (dark box). We solve a more general learning problem (light box) by jointly optimizing the dataset $\cD_{\alpha}$ in addition to $\wb$. To avoid the trivial solution $\alpha = 0$, it is customary in AutoML to optimize $\cD_\alpha$ by minimizing the validation error computed on a disjoint dataset. This makes for inefficient use of the data, which has to be split between training and validation sets. Instead, we leverage a closed-form expression of the leave-one-out cross-validation error to jointly optimize the model and data weights during training, without the need to create a separate validation set. 
\begin{center}
    \includegraphics[width=0.6\linewidth]{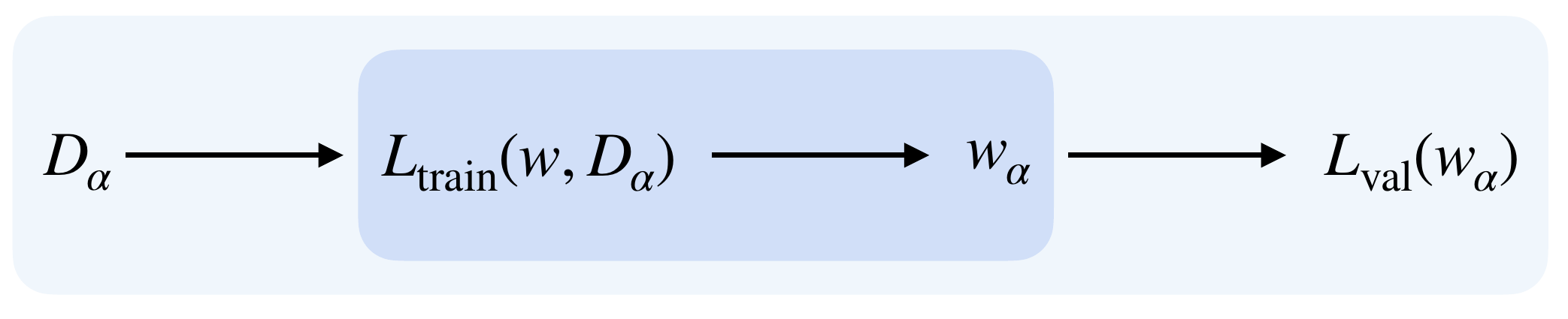}
\end{center}
The intermediate block in the diagram (which finds the optimal weights $\wb_{\alpha}$ for the the training loss on $\cD_{\alpha}$) is usually non-differentiable with respect to the dataset, or the derivative is prohibitively expensive to compute. DIVA leverages recent progress in deep learning linearization \cite{achille2020lqf}, to derive a closed-form expression for the derivative of the final loss (validation error) with respect to the dataset weights. In particular, \cite{achille2020lqf} have shown that, by replacing cross-entropy with least-squares, replacing ReLu with leaky-ReLu, and performing suitable pre-conditioning, the linearized model performs on par with full non-linear fine-tuning. We also leverage a classical result to compute the leave-one-out loss of a linear model in closed-form \citep{rifkin2007notes,green1993nonparametric}. This allows us to optimize the LOO loss without requiring a separate validation set, setting DIVA apart from  bi-level optimization customary in AutoML. 

To illustrate the many possible uses of the dataset derivative, we run experiments with a simplified version of DIVA to cleanup a dataset of noisy annotations, to extend a training set with additional data from an external pool, identify meaningful data augmentation, and to perform multi-modal expansion using a CLIP model \citep{radford2021learning}. 

Rather than using the full linearization of the model derived by \cite{achille2020lqf}, we restrict the gradient to its last layer, cognizant that we are not exploiting the full power of LQF and thereby obtaining only a lower-bound of performance improvement. Despite that restriction, our results show consistent improvements from dataset optimization, at the modest computational cost of a forward pass over the dataset to optimize the importance weights.

To summarize, our main contributions are:
\begin{enumerate}
    \item We introduce a method to compute the dataset derivative in closed form, DIVA.
    \item We illustrate the use of DIVA to perform dataset optimization by minimizing directly the leave-one-out error without the need for an explicit validation dataset.
    \item We perform experiments with a simplified model that, despite not using the full power of the linearization, shows consistent improvements in dataset extension, re-weighting, outlier rejection and automatic aggregation of multi-modal data.
\end{enumerate}

Our method presents several limitations. The dataset derivative of a learning task is computed around a point represented by a pre-trained model. It only allows local optimization around this point.  Moreover, we only compute a restriction of the linearization to the dimensions spanned by the last few layers. In general, this yields suboptimal results compared to full global optimization from scratch, if one could compute that at scale. Nonetheless, the linearized setting is consistent with the practice of fine-tuning pre-trained models in light of the results of \cite{achille2020lqf}, see also \citep{radford2021learning, rezende2017malicious,mormont2018comparison,hosny2018skin}.

\section{Related Work}

\paragraph{AutoML.}
State of the art performance in image classification tasks often relies on large amount of human expertise in selecting models and adjusting the training settings for the task at hand \citep{li2020rethinking}. Automatic machine learning (AutoML) \citep{feurer2019auto, he2021automl} aims to automate model selection \citep{cawley2010over} and the training settings by instead using meta-algorithms for the different aspects of the learning settings. Such methods follow a bi-level optimization framework, optimizing the training settings in the outer level, and traditional model optimization in the inner level \citep{jenni2018deep}. AutoML has focused on achieving better results via automatic model selection \citep{deshpande2021linearized, feurer2019auto} including neural architecture search (NAS) \citep{zoph2016neural,elsken2019neural, liu2019auto}. Other important AutoML topics include hyper-parameter selection \citep{li2017hyperband, akiba2019optuna} and data augmentation \citep{cubuk2018autoaugment,lim2019fast, chen2020hypernetwork, behl2020autosimulate}, which are closer to our settings of optimizing the dataset weights. Since the main signal for a model's performance is the final validation loss, which requires full optimization of the model for each evaluation, AutoML approaches often incur a steep  computational costs. Alternatively, other methods follow alternating optimization of the criteria, such as the work of \cite{ren2018learning} that approximates full network optimization with a single SGD step to learn to reweight the training set dynamically.
\emph{Differentiable AutoML} alleviates outer-optimization costs while optimizing the final validation error via differentiable programming, by utilizing proxy losses and continuous relaxation that enable differentiation. 
Different approaches to differentiable AutoML include differentiable NAS \citep{liu2018darts, wu2019fbnet}, data augmentation \citep{liu2021direct, li2020dada}, and hyper-parameter optimization \citep{andrychowicz2016learning}. The DIVA dataset derivative follows the differentiable AutoML framework by enabling direct optimization of the dataset with respect to the final validation error of the model.  

\paragraph{Importance sampling.} While our dataset optimization problem may seem superficially similar to importance sampling, the optimization objective is different. Importance sampling aims to reweight the training set to make it more similar to the test distribution or to speed up convergence. On the other hand, DIVA objective is to optimizes a validation loss of the model, even if this requires making the training distribution significantly different from the testing distribution. Importance sampling methods have a long history in the MCMC machine learning literature where the sampling is conditioned on the predicted importance of samples \citep{metropolis1949monte, liu2008monte}. In deep learning, importance sampling methods have been studied theoretically for linearly-separable data \citep{byrd2019effect} and recently in more generality \citep{xu2021understanding}. Furthermore, there exist many importance sampling heuristics in deep learning training including different forms of hard sample mining \citep{shrivastava2016training, xue2019hard, chang2017active}, weighting based on a focal loss \citep{lin2017focal}, re-weighting for imbalance, \citep{cui2019class, huang2019deep, dong2017class} and gradient based scoring \citep{li2019gradient}. We emphasize that DIVA's optimization of the sample weights is not based on a heuristic but is rather a differentiable AutoML method driven by optimization of a proxy of the test error. Further, DIVA allows optimization of the dataset weights with respect to an arbitrary loss and also allows for dataset extension computation. 

\paragraph{LOO based optimization.} Leave-one-out cross validation is well established in statistical learning \citep{stone1977asymptotics}. In ridge regression, the LOO model predictions for the validation samples have a closed-form expression that avoids explicit cross validation computation \citep{green1993nonparametric,rifkin2007notes} enabling efficient and scalable unbiased estimate of the test error. Efficient LOO has been widely used as a criterion for regularization \citep{pedregosa2011scikit,quan2010weighted,birattari1999lazy,thapa2020adaptive}, hyper-parameter selection \citep{hwang2017geographically} and optimization
\citep{wen2008heuristic}. Most similar to our dataset derivative are methods that: (1) optimize a restricted set of parameters, such as kernel bandwidth, in weighted least squares \citep{cawley2006leave,hong2007kernel} (2) locally weighted regression methods (\emph{memorizing regression}) \citep{atkeson1997locally,Moore92anempirical}, or (3) methods that measure the impact of samples based on LOO predictions \citep{brodley1999identifying,nikolova2021outlier}.

\paragraph{Dataset selection \& sample impact measures.}
\cite{koh2017understanding} measure the effect of changes of a training sample weight on a final validation loss through per-sample weight gradients, albeit without optimizing the dataset and requiring a separate validation set.
Their proposed expression for the per-sample gradient, however, does not scale easily to our problem of dataset optimization. In contrast, in \cref{prop:loo_derivative} we introduce an efficient closed-form expression for the derivative of the whole datasets. Moreover, in \cref{prop:loo_derivative}, we show how to optimize the weights with respect to a cross-validation loss which does not require a separate set.
In \cite{pruthi2020estimating}, the authors present a sample-impact measure for interpretability based on a validation set; for dataset extension, \cite{yan2020neural} presents a coarse dataset extension method based on self-supervised learning. Dataset distillation and core set selection methods aim to decrease the size of the dataset \citep{wang2018dataset} by selecting a representative dataset subset \citep{hwang2020data,jeong2020dataset,coleman2019selection,joneidi2020select,trichet2018dataset, DBLP:journals/corr/abs-2103-00123}. While DIVA is capable of removing outliers, in this work we do not approach dataset selection from the perspective of making the dataset more computationally tractable by reducing the number of samples.

\section{Method}\label{sec:method}
In supervised learning, we use a parametrized model $f_\wb(\xb)$ to predict a target output $y$ given an input $\xb$ coming from a joint distribution $(\xb,y) \sim \cT$. Usually, we are given a training set $\cD = \{(\xb_i, y_i)\}_{i=1}^N$ with samples $(x, y)$ assumed to be independent and identically distributed (i.i.d.) according to $\cT$. The training set $\cD$ is then used to assemble the empirical risk for some per-sample loss $\ell$, 
\begin{equation*}
    L_\text{train}(\wb; \cD) = \sum_{i=1}^N \ell(f_\wb(\xb_i), y_i),
\end{equation*}  which is minimized to find the optimal model parameters $\wb_\cD$:
\begin{equation*}
    \wb_\cD = \argmin_\wb L_\text{train}(\wb; \cD).
\end{equation*}
The end goal of empirical risk minimization is that weights will also minimize the {\em test loss}, computed using a separate test set. Nonetheless $\cD$ is often biased and differs from the distribution $\cT$. In addition, from the perspective of optimization, different weighting of the training loss samples can enable or inhibit good learning outcomes of the task $\cT$ \citep{lin2017focal}.

\paragraph{Dataset  Optimization.} In particular, it may not be the case that sampling the training set $\cD$ i.i.d.\ from $\cT$ is the best option to guarantee generalization, nor it is realistic to assume that $\cD$ is a fair sample. Including in-distribution samples that are too difficult may negatively impact the optimization, while including certain out-of-distribution examples may aid the generalization on $\cT$. It is not uncommon, for example, to improve generalization by training on a larger dataset containing out-of-distribution samples coming from other sources, or generating out-of-distribution samples with data augmentation. We call Dataset Optimization the problem of finding the optimal subset of samples, real or synthetic, to include or exclude from a training set $\cD$ in order to guarantee that the weights $\wb_\cD$ trained on $\cD$ will generalize as much as possible.

\paragraph{Differentiable Dataset Optimization.}
Unfortunately, a na\"ive brute-force search over the $2^N$ possible subsets of $\cD$ is unfeasible. The starting idea of DIVA is to instead solve a more general continuous optimization problem that can however be optimized end-to-end.

Specifically, we parameterize the choice of samples in the augmented dataset through a set of non-negative continuous sample weights $\alpha_i$ which can be optimized by gradient descent along with the weights of the model. Let $\alpha=(\alpha_1, \ldots, \alpha_N)$ be the vector of the sample weights and denote the corresponding weighted dataset by $\cD_\alpha$. The training loss on $\cD_\alpha$ is then defined as:
\begin{equation}
\label{eq:weighted_loss}
L_\train(\wb; \cD_\alpha)= \sum_{i=1}^N \alpha_i\, \ell(f_\wb(\xb_i), y_i).
\end{equation}
Note that if all $\alpha_i$'s are either 0 or 1, we are effectively selecting only a subset of $\cD$ for training. As we will show, this continuous generalization allows us to optimize the sample selection in a differentiable way.
In principle, we would like to find the sample weights $\alpha^*= \argmin_\alpha L_\test(\wb_{\alpha})$ that lead to the best generalization. Since we do not have access to the test data, in practice this translates to optimizing $\alpha$ with respect to an (unweighted) validation loss $L_\val$:
\[
\alpha^* = \textstyle \argmin_\alpha L_\val(\wb_{\alpha}).
\]
We can, of course, compute a validation loss using a separate validation set. However, as we will see in Section \ref{sec:loo_derivation}, we can also use a leave-one-out cross-validation loss directly on the training set, without any requirement of a separate validation set.

In order to efficiently optimize $\alpha$ by gradient-descent, we need to compute the dataset derivative $\nabla_{\alpha} L_{\val}(\wb_{\alpha})$.
By the chain rule, this can be done by computing $\nabla_\alpha \wb_{\alpha}$. However, the training function $\alpha \to \wb_{\alpha}$ that finds the optimal weights $\wb_{\alpha}$ of the model given the sample weights $\alpha$ 
may be non-trivial to differentiate or may not be differentiable at all (for example, it may consist of thousands of steps of SGD). This would prevent us from minimizing $\alpha$ end-to-end.

In the next section, we show that if, instead of linearizing the $\wb_\alpha$ end-to-end in order to compute the derivative, we linearize the model \textit{before} the optimization step, the derivative can both be written in closed-form and computed efficiently, thus giving us a tractable way to optimize $\alpha$.

\begin{figure}[H]
  \begin{center}
    \includegraphics[width=0.6\linewidth]{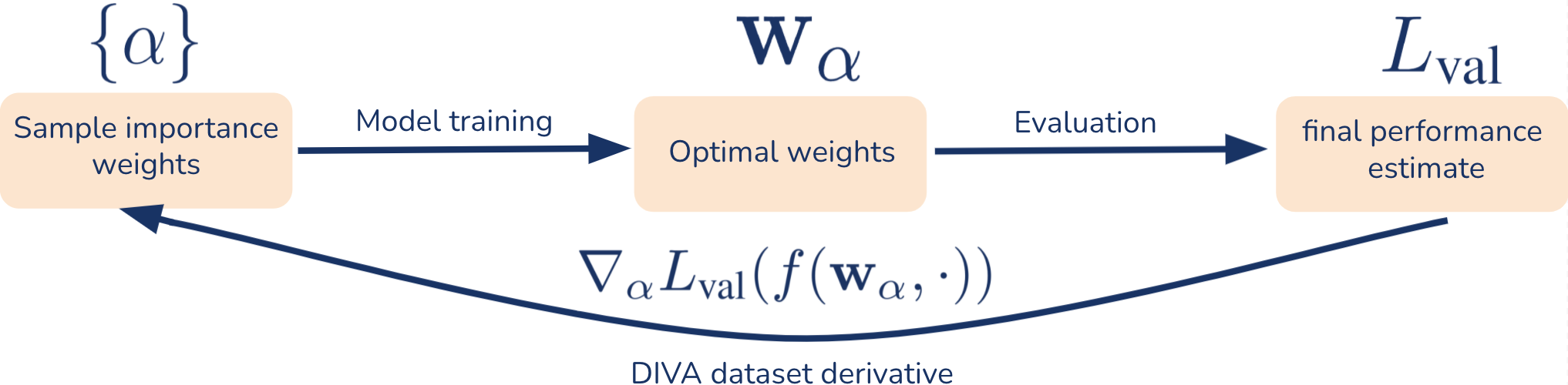}
\end{center}
\caption{The DIVA dataset derivative is computed end-to-end from the final validation loss}
    \label{fig:diva_diagram}
\end{figure}
\subsection{Linearization} \label{sec:linearization}

In real-world applications, the parametric model $f_\wb(\xb)$ is usually a deep neural network.
Recent work \citep{achille2020lqf,mu2020gradients} have shown that in many cases, a deep neural network can be transformed to an equivalent linear model that can be trained on a simple quadratic loss and still reach a performance similar to the original model. Given a model $f_\wb(\xb)$, let $\wb_0$ denote an initial set of weights, For example, $\wb_0$ could be obtained by pre-training on a large dataset such as ImageNet (if the task is image classification). Following \cite{achille2020lqf,mu2020gradients}, we consider a linearization $f^\lin_\wb (\xb)$ of the network $f_\wb(\xb)$ given by the first-order Taylor expansion of $f_\wb(\xb)$ around $\wb_0$:
\begin{equation}
f^\lin_\wb(\xb) = f_{\wb_0}(\xb) + \nabla_\wb f_{\wb_0}(\xb) \cdot (\wb - \wb_0).
\end{equation}
Intuitively, if fine-tuning does not move the weights much from the initial pre-trained weights $\wb_0$, then $f^\lin_\wb(\xb)$ will remain a good approximation of the network while becoming linear in $\wb$ (but still remaining highly non-linear with respect to the input $\xb$). Effectively, this is equivalent to training a linear classifier using the gradients $\zb_i := \nabla_\wb f_{\wb_0}(\xb_i)$ as features \citep{mu2020gradients}.
Although $f^\lin_\wb(\xb)$ is a linear model, the optimal weights $\wb_\alpha$ may still be a complex function of the training data, depending on the loss function used. \cite{achille2020lqf}  showed that equivalent performance can be obtained by replacing the empirical cross-entropy with the regularized least-squares loss:
\begin{equation}
L_\train(\wb) = \sum_{i=1}^N \|f^\lin_\wb(\xb) - \yb_i\|^2 + \lambda \|\wb\|^2
\label{eq:unweighted_l2_loss}
\end{equation}
where $\yb$ denotes the one-hot encoding vector of the label $y_i$. In \cite{achille2020lqf}, it is shown that linearized models are equivalent from the standpoint of performance on most standard tasks and classification benchmarks, and better in the low-data regime, which is where the problem of ``dataset augmentation'' is most relevant. The advantage of using this loss is that the optimal weights $\wb^*$ can now be written in closed-form as
\begin{equation}
\label{eq:l2-solution}
\wb^* = (\Zb^\top \Zb + \lambda \Ib)^{-1} \Zb^\top (\Yb - f_{\wb_0}(\Xb)),
\end{equation}
where $\Zb = [\zb_1, \ldots, \zb_N]$ is the matrix of the Jacobians $\zb_i = \nabla_\wb f_{\wb_0}(\xb_i)$. While our method can be applied with no changes to linearization of the full network, for simplicity in our experiments we restrict to linearizing only the last layer of the network.
This is equivalent to using the network as a fixed feature extractor and training a linear classifier on top the last-layer features, that is, $\zb_i = f_{\wb_0}^{L-1}(\xb_i)$ are the features at the penultimate layer.

\subsection{Computation of the Dataset Derivative}
We now show that for linearized models we can compute the derivative $\nabla_\alpha \wb_{\alpha}$ in closed-form.
For the $\alpha$-weighted dataset, the objective in  \cref{eq:unweighted_l2_loss} with $L_2$ loss for the linearized model is written as,
\begin{equation}
   \wb_{\alpha} = \argmin_{\wb} L_{\train}(\wb; \cD_{\alpha}) = \argmin_{\wb} \sum_{i=1}^N \alpha_i\|\wb^\top\zb_i - \yb_i\|^2 + \lambda\|\wb\|^2.
   \label{eq:weighted_l2_ridge}
\end{equation}
where $\zb_i = \nabla_\wb f_{\wb_0}(\xb_i)$ as in the previous section. Note that $\alpha_i \|\wb^\top \zb_i - \yb_i\|^2 = \|\wb^\top \zb_i^\alpha - \yb_i^\alpha\|$, where $\zb^\alpha_i := \sqrt{\alpha_i} \zb_i$ and $\yb^\alpha_i := \sqrt{\alpha_i} \yb_i$. Using this, we can reuse \cref{eq:l2-solution} 
to obtain the following closed-form solution for $\wb_\alpha$:
\begin{equation}
     \wb_\alpha = (\Zb^\top \Db_\alpha \Zb + \lambda \Ib)^{-1}\Zb^\top \Db_\alpha \Yb,
     \label{eq:closed_form_sol}
\end{equation}
where we have taken $\Db_\alpha = \diag(\alpha)$.
In particular, note that $\wb_\alpha$ is now a differentiable function of $\alpha$. The following proposition gives a closed-form expression for the derivative.
\begin{proposition}
[Model-Dataset Derivative $\nabla_\alpha \wb_\alpha$]\label{prop:model_dataset} For the ridge regression problem \eqref{eq:weighted_l2_ridge} and $\wb_{\alpha}$ defined as in \eqref{eq:closed_form_sol}, define 
\begin{equation}
\Cb_\alpha = (\Zb^\top \Db_\alpha \Zb + \lambda \Ib)^{-1}.
\end{equation}
Then the Jacobian of $\wb_{\alpha}$ with respect to $\alpha$ is given by
\begin{equation}
    \nabla_{\alpha}\wb_{\alpha} = \Zb \Cb_\alpha \circ \big((\Ib - \Zb \Cb_\alpha \Zb^\top\Db_\alpha)\Yb\big),
\end{equation}
\end{proposition}
Where we write $\Ab \circ \Bb \in \RR^{n \times m \times k}$ for the batch-wise outer product of $\Ab\in \RR^{n\times m}$ and $\Bb\in \RR^{n\times k}$ along the common dimension $k$, i.e., $(\Ab \circ \Bb)_{ijk} = a_{ij} b_{ik}$.

The Jacobian $\nabla_\alpha \wb_{\alpha}$ would be rather large to compute explicitly. Fortunately, the end-to-end gradient of the final validation loss, $L_{\val}(\wb_{\alpha})$, can still be computed efficiently, as we now show.  Given a validation dataset $\cD_\val$, the validation loss is:
\begin{equation}
    \textstyle L_{\val}(\wb_{\alpha}) = \sum_{(\xb_i, y_i) \in \cD_\val} \ell(f_{\wb_{\alpha}}(\xb_i), y_i).
\end{equation}
The following gives the expression from which we optimize $\alpha$ end-to-end with respect to the validation loss.
\begin{proposition}[Validation Loss Dataset Derivative]\label{prop:val_dataset}
Define $\Lb$ as the loss function derivative with respect to the network outputs as, 
\begin{equation*}
    \Lb = \bigg[\pdv{\ell}{f}(f(\xb_1), y_1), \cdots \pdv{\ell}{f}(f(\xb_N), y_N)\bigg]
\end{equation*}
Then the dataset derivative importance weights with respect to final validation is given by
\begin{equation}
    \nabla_{\alpha} L_{\val}(\wb_{\alpha}) = \Zb\Cb_\alpha \Zb^\top \times\big( \Lb^\top \Yb^\top (\Ib - \Db_{\alpha}\Zb \Cb_\alpha \Zb^\top)\big).
\end{equation}
\end{proposition}

\begin{figure}
    \centering
    \vspace*{-20mm}
    \includegraphics[width=0.49\linewidth]{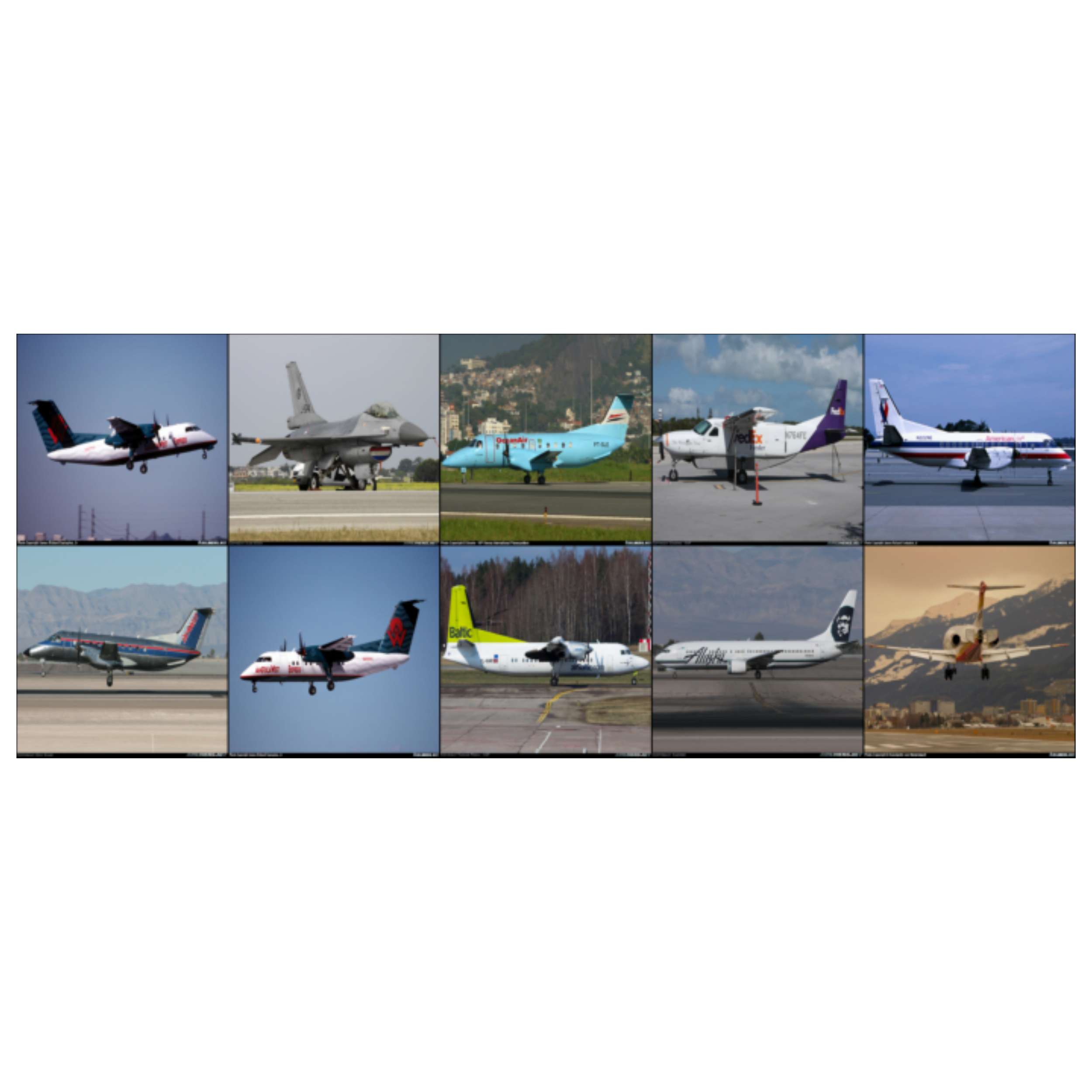}
    \includegraphics[width=0.49\linewidth]{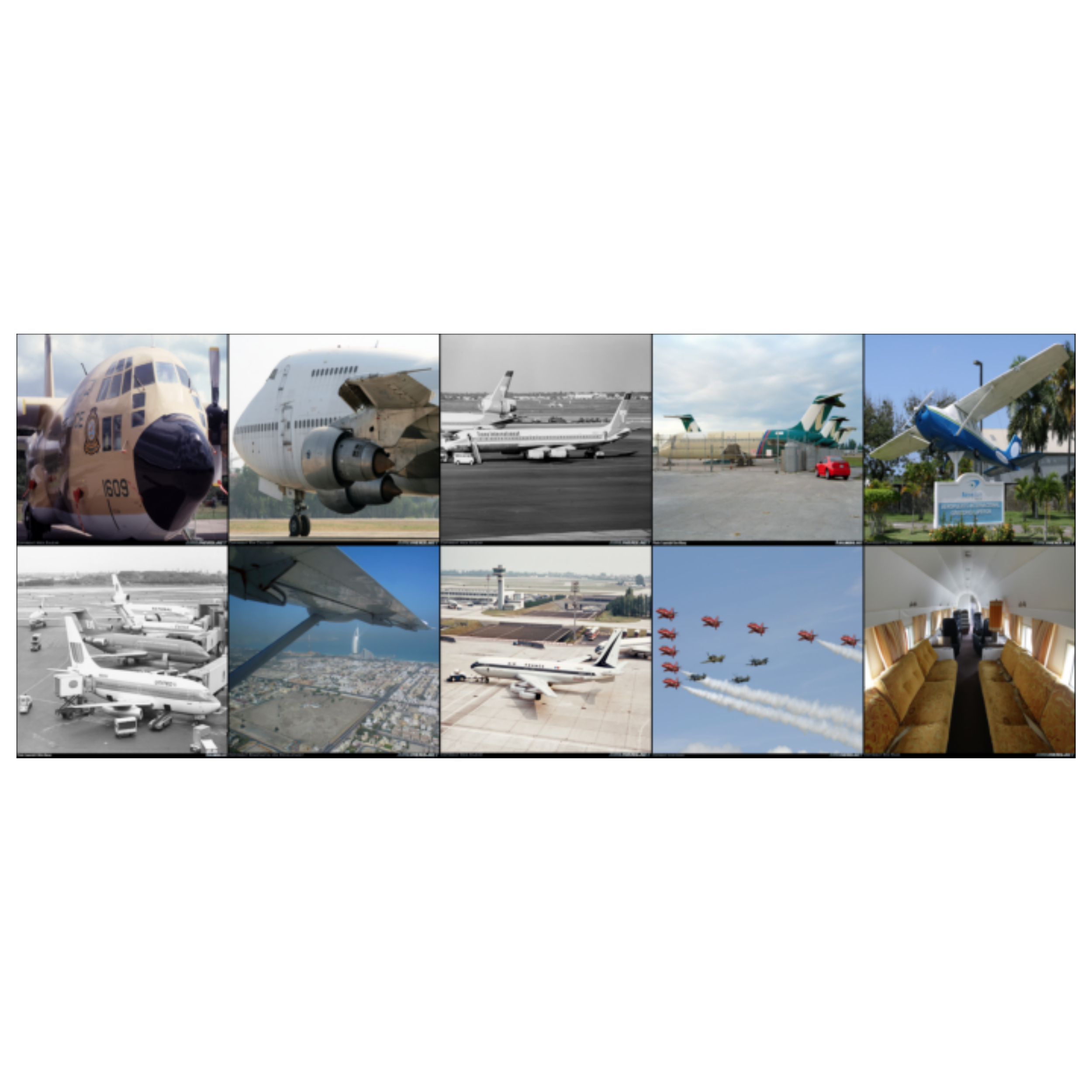}
    \vspace*{-22mm}
    \caption{\textbf{Examples of the reweighting done by DIVA.} \textbf{(Left)} Samples from the FGVC Aircraft classification dataset that are up-weighted by DIVA and \textbf{(Right)} samples that are down-weighted because they increase the test error. Down-weighted samples tend to have planes in non-canonical poses, multiple planes, or not enough information to classify the plane correctly.}
    \label{fig:qualitative-airplanes}
\end{figure}

\subsection{Leave-one-out optimization}\label{sec:loo_derivation}

It is common in AutoML to optimize the hyper-parameters with respect to a separate validation set. However, using a separate validation may not be practical in limited data settings, which are a main focus of dataset optimization. To remedy this, we now show that
we can instead  optimize $\alpha$ by minimizing a leave-one-out cross-validation loss that only requires a training set:
\begin{equation}
    \textstyle L_{\loo}(\alpha) = \sum_{i=1}^N \ell(f_{\wb_\alpha^{-i}}(\xb_i), y_i),
    \label{eq:loo}
\end{equation}
where $\wb^{-i}_\alpha$ are the optimal weights obtained by training with the loss \cref{eq:weighted_l2_ridge} on the entire dataset ${\cal D}$ {\em except} for the $i$-th sample $(\xb_i, y_i)$.
This may seem counter-intuitive, since we are optimizing the weights of the training samples using a validation loss defined on the training set itself. It is useful to recall that $\wb_\alpha^{-i}$ minimizes the $\alpha$-weighted $L_2$ loss on the training set (minus the $i$-th example):
\begin{equation}
  \wb_\alpha^{-i} = \arg\min_\wb L_\train^{-i}(w, \cD_\alpha) =  \arg\min_\wb \sum_{j \neq i} \alpha_j \|f_w(x_j) - y_j\|^2 + \lambda \|\wb\|^2.
  \label{eq:weighted_loo_dataset}
\end{equation}
Meanwhile, $\alpha$ minimizes the \textit{unweighted} validation loss in \cref{eq:loo}. This prevents the existence of degenerate solutions for $\alpha$.

Computing $L_{\loo}$ naively would require training $n$ classifiers, but
fortunately, in the case of a linear classifier
with the $L_2$ loss, a more efficient closed-form solution exists \citep{green1993nonparametric,rifkin2007notes}. Generalizing those results to the case of a weighted loss, we are able to derive the following expression.
\begin{proposition}\label{prop:loo_derivative}
Define \begin{equation*}
\Rb_{\alpha} = \Zb^\top \sqrt{\Db_{
\alpha}}(\Zb^\top \Db_{\alpha} \Zb + \lambda \Ib)^{-1}\sqrt{\Db_{\alpha}}\Zb
\end{equation*}
Then $\alpha$-weighted LOOV predictions defined in \cref{eq:weighted_loo_dataset} admit a closed-form solution:
\begin{equation}
    f_{\wb_{\alpha}^{-i}}(\zb_i) = \Bigg[ \frac{\Rb_{\alpha}\sqrt{\Db_{\alpha}}\Yb - \diag(\Rb_{\alpha})\sqrt{\Db_{\alpha}}\Yb}{\diag(\sqrt{\Db_{\alpha}} - \sqrt{\Db_{\alpha}}\Rb_{\alpha})}\bigg]_i,
    \label{eq:weighted_loo}
\end{equation}
where $\diag(\Ab)=[a_{11},\ldots,a_{nn}]$ denotes the vector containing the diagonal of $\Ab$, and the division between vectors is element-wise.
\end{proposition}

Note that the prediction $f_{\wb_{\alpha}^{-i}}(\zb_i)$ on the $i$-th sample when training on all the other samples is a differentiable function of $\alpha$.
Composing \cref{eq:weighted_loo} in \cref{eq:loo}, we compute the derivative $\nabla_\alpha L_{\loo}(\alpha)$, which  allows us to optimize the cross-validation loss with respect to the sample weights, without the need of a separate validation set. We give the closed-form expression for $\nabla_\alpha L_{\loo}(\alpha)$ in the Appendix.

\subsection{Dataset Optimization with DIVA}
\label{sec:diva-applications}

We can now apply the closed-form expressions for $\nabla_\alpha L_\val(\alpha)$ and $\nabla_\alpha L_\loo(\alpha)$ for differentiable dataset optimization.  We describe the optimization using $L_\val$, but the same applies to $L_\loo$.

\paragraph{DIVA Reweight.}
The basic task consists in reweighting the samples of an existing dataset in order to improve generalization. This can curate a dataset by reducing the influence of outliers or wrong labels, or by reducing possible imbalances. 
To optimize the dataset weights, we use gradient descent in the form:
\begin{equation}
\label{eq:alpha-gradient-descent}
    \alpha \leftarrow \alpha - \eta \nabla_{\alpha}L_{\val}.
\end{equation}
It is important to notice that $L_\val$ is an unbiased estimator of the test loss only at the first step, hence optimizing using \cref{eq:alpha-gradient-descent}
for multiple steps can lead to over-fitting (see Appendix). Therefore, we apply only 1-3 gradient optimization steps with a relatively large learning rate $\eta \simeq 0.1$. This early stopping both regularizes the solution and decreases the wall-clock time required by the method. We initialize $\alpha$ so that $\alpha_i=1$ for all samples.

\paragraph{DIVA Extend.}
The dataset gradient also allows us to extend an existing dataset. Given a core dataset $\cD=\{(\xb_i, y_i)\}_{i=1}^{N}$ and an external (potentially noisy) data pool $\cE=\{(\hat{\xb}_i, \hat{y}_i)\}_{i=N+1}^{N+M}$, we want to find the best samples from $\cE$ to add to $\cD$. For this we merge $\cD$ and $\cE$ in a single dataset and initialize $\alpha$ such that $\alpha_i=1$ for samples of $\cD$ and $\alpha_i=0$ for samples of $\cE$ (so that initially the weighted dataset matches $\cD$). We then compute $\nabla_\alpha L_\val(\alpha)$ to find the top $k$ samples of $\cE$ that have the largest negative value of $\nabla_\alpha L_\val(\alpha)_i$, i.e., the samples that would give the largest reduction in validation error if added to the training set and add them to $\cD$. This is repeated until the remaining samples in $\cE$ all have positive value for the derivative (adding them would not further improve the  performance).

\paragraph{Detrimental sample detection.} The $i$-th component of $\nabla_{\alpha}L_{\val}$ specifies the influence of the $i$-th sample on the validation loss. In particular, 
$(\nabla_{\alpha}L_{\val})_i > 0$ implies that the sample increases the validation loss, hence it is detrimental (e.g., it is mislabeled or overly represented in the dataset). We can select the set of detrimental examples by thresholding $\nabla_{\alpha}L_{\val}$:
\begin{equation}
    \text{Detrimental}(\epsilon) = \big\{i :~ (\nabla_{\alpha}L_{\val})_i \ge \epsilon \big\}.
\end{equation}

\section{Results}
For our models we use standard residual architectures (ResNet) models pre-trained on  ImageNet \citep{deng2009imagenet} and Places365 \citep{zhou2017places}.
For our experiments on dataset optimization
we consider datasets that are smaller than the large scale datasets used for pre-training as we believe they reflect more realistic conditions for dataset optimization. For our experiments we use the CUB-200 \citep{WelinderEtal2010}, FGVC-Aircraft, \citep{maji13fine-grained}, Stanford Cars \citep{KrauseStarkDengFei-Fei_3DRR2013}, Caltech-256 \citep{griffin2007caltech}, Oxford Flowers 102 \citep{nilsback2008automated}, MIT-67 Indoor \citep{quattoni2009recognizing}, Street View House Number \citep{netzer2011reading}, and the Oxford Pets \citep{parkhi2012cats} visual recognition and classification datasets. In all experiments, we use the network as a fixed feature extractor, and train a linear classifier on top of the network features using the weighted $L_2$ loss \cref{eq:weighted_l2_ridge} and optimize the $\alpha$ weights using DIVA.
\begin{table}
\centering
{\small
\begin{tabular}{rcccccc}\toprule
 Dataset & Original & DIVA Reweight & {(Chang et al.) \cite{chang2017active}}  & {(Ren et. al) \cite{ren2018learning}} & Gain \\
\midrule
\midrule
Aircrafts &	57.58&	\textbf{54.64}&  70.48& 81.82 (80.62)&	+2.94\\
Cub-200 &	39.30&	\textbf{36.93}& 57.85 & 72.55 (75.35) &	+2.36 \\ 
MIT Indoor-67  &	32.54&	\textbf{31.27}& 37.84 & 64.48 (58.06) &	+1.27 \\
Oxford Flowers &	20.23 &	\textbf{19.16} &	22.82 & 48.80 (55.46) & +1.07 \\
Stanford Cars &	58.91&	\textbf{56.31}& 75.87 & 83.09 (84.50) &	+2.56	 \\
Caltech-256	& 23.98& \textbf{21.29}& 37.52 & 58.44 (52.77) &	+2.69	 \\
\bottomrule
\end{tabular}
}

\vspace{.5em}
\caption{\label{tab:diva-reweigh-clean}Test error of DIVA Reweight to curate several fine-grain classification datasets. We use a ResNet-34 pretrained on ImageNet as feature extractor and train a linear classifier on top of the last layer. Note that DIVA Reweight can improve performance even on curated and noiseless datasets whereas other reweighting methods based on hard-coded rules may be detrimental in this case.
}
\end{table}

\paragraph{Dataset AutoCuration.} We use DIVA Reweight to optimize the importance weights of samples from several fine-grain classification datasets. While the datasets have already been manually curated by experts to exclude out-of-distribution or mislabeled examples, we still observe that in all cases DIVA can further improve the test error of the model (\Cref{tab:diva-reweigh-clean}). 
To understand how DIVA achieves this, in \Cref{fig:qualitative-airplanes}
we show the most up-weighted (left) and down-weighted (right) examples on the FGVC Aircraft classification task \cite{maji13fine-grained}. We observe that DIVA tends to give more weight to clear, canonical examples, while it detects as detrimental (and hence down-weights) examples that contain multiple planes (making the label uncertain), or that do not clearly show the plane, or show non-canonical poses. We compare DIVA Reweight with two other re-weighting approaches: \cite{ren2018learning}, that applies re-weighting using information extracted from a separate validation gradient step, and \cite{chang2017active}, which reweighs based on the uncertainty of each prediction (threshold-closeness weighting scheme). For \cite{ren2018learning},  we set aside 20\% of the training samples as validation for the reweight step, but use all samples for the final training (in parentheses). We notice that both baselines, which are designed to reweight noisy datasets, underperform with respect to DIVA on datasets without artificial noise.

\begin{figure}[H]
    \centering
     \includegraphics[width=0.30\linewidth]{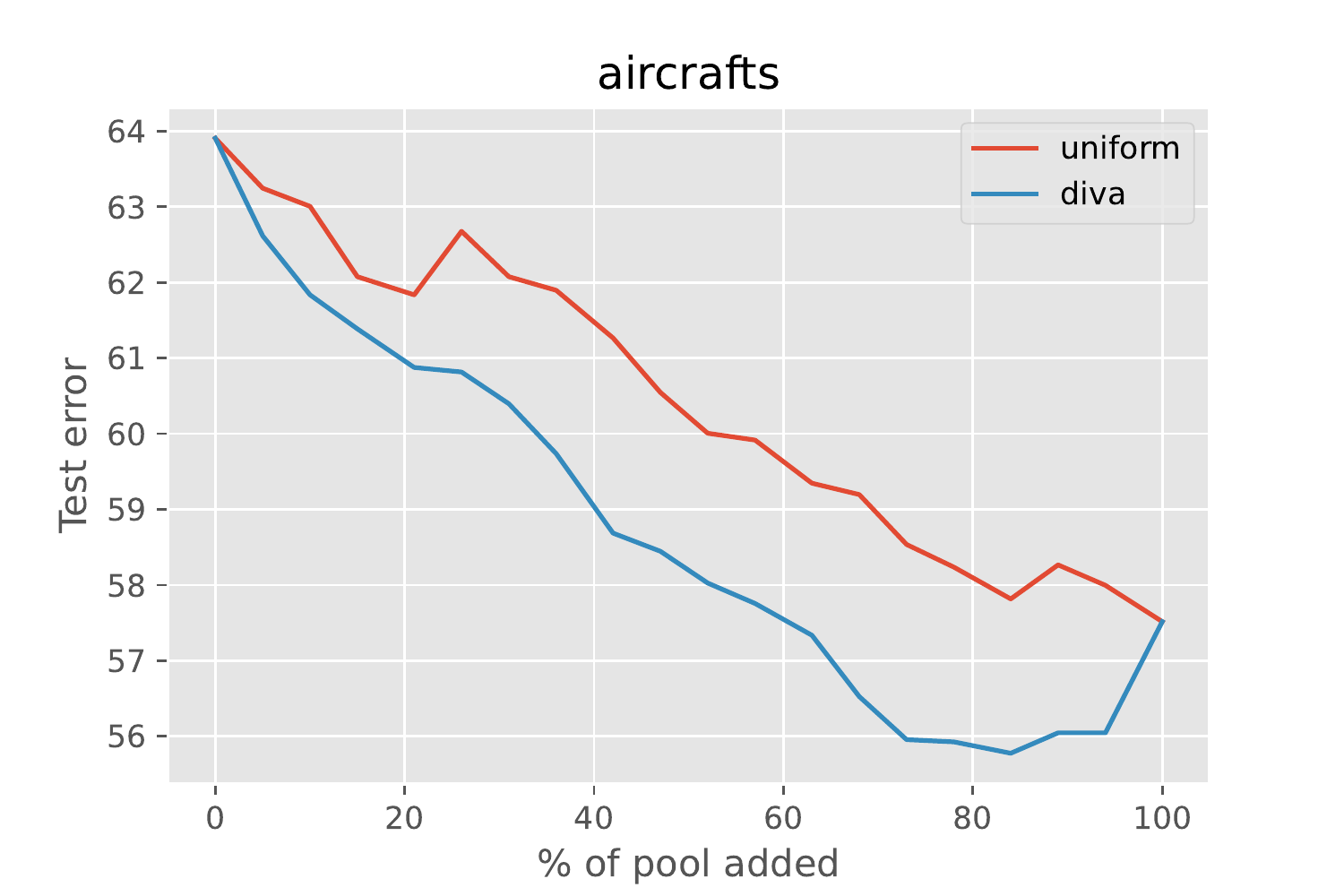} \includegraphics[width=0.30\linewidth]{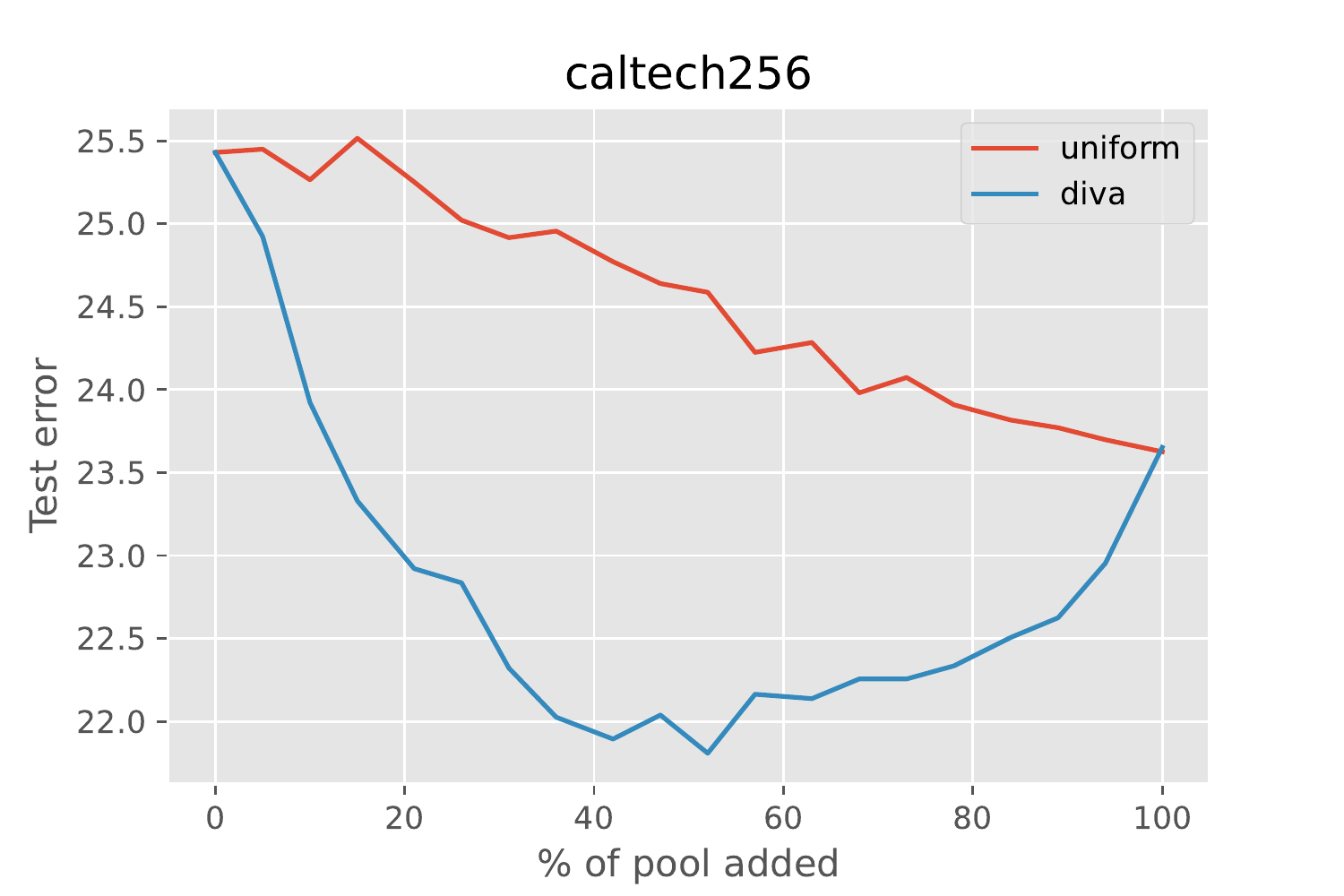}
     \includegraphics[width=0.30\linewidth]{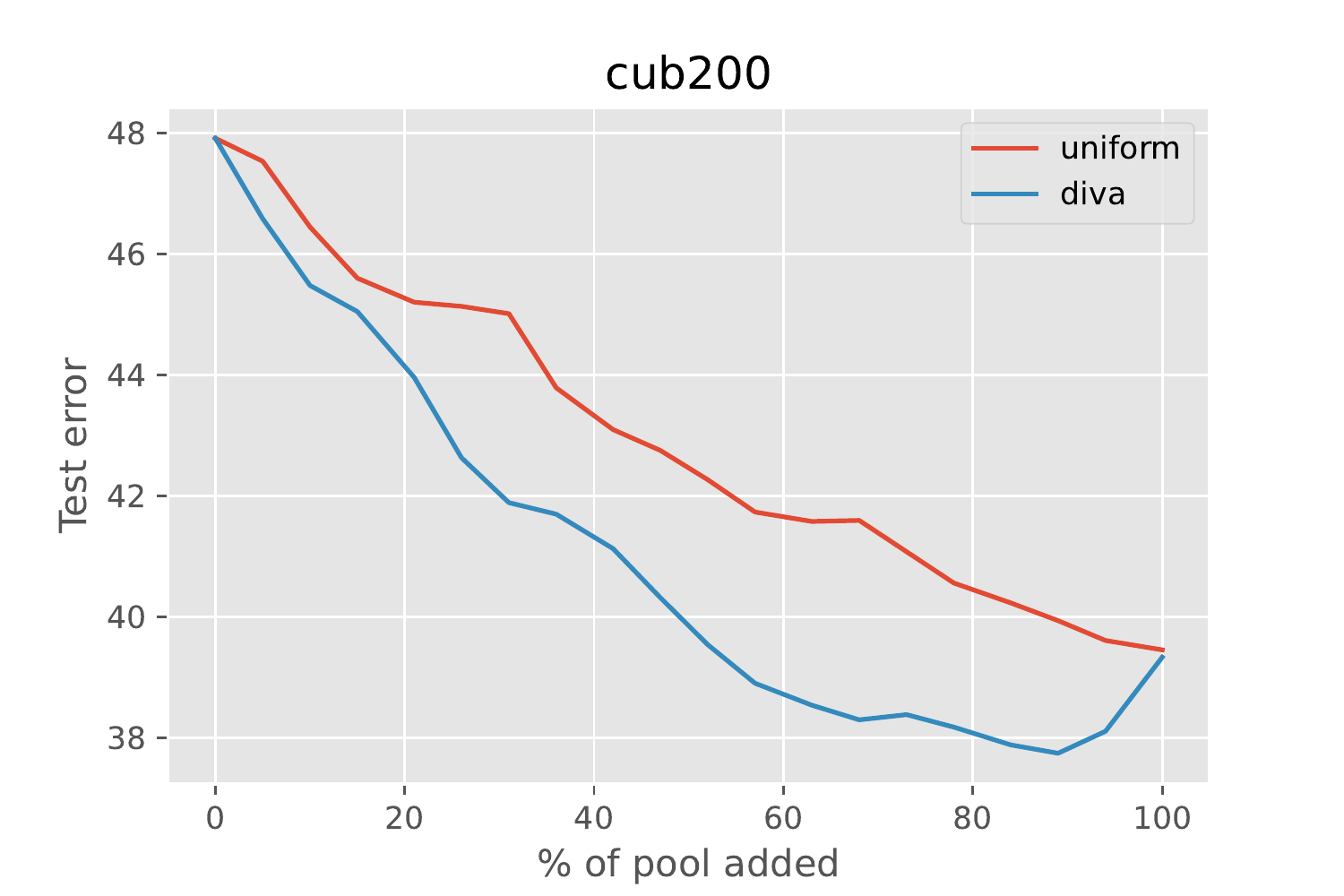}
    \caption{\textbf{DIVA Extend.} We show the test error achieved by the model as we extend a dataset with samples selected from a dataset pool using either DIVA Extend (red line) or uniform sampling (blue line). The pool set matches the same distribution as the training set. In all cases DIVA Extend outperforms uniform sampling and identifies subsets of the pool set with better performance than the whole pool. We also note that using only a subset selected by DIVA as opposed to using the whole pool, actually improves the test accuracy.
    }
    
    \vspace{-1.2em}
    
    \label{fig:extend_increment}
\end{figure}

\paragraph{Dataset extension.} We test the capabilities of DIVA Extend to extend a dataset with additional samples of the distribution. In \Cref{fig:extend_increment} and \Cref{tab:diva-extend}, we observe that DIVA is able to select the most useful examples and reaches an optimal performance generalization error using significantly less samples than the baseline uniform selection. Moreover, we notice that DIVA identifies a smaller subset of samples that provides better test accuracy than adding all the pool samples to the training set.

\begin{table}[H]
\centering
\resizebox{.55\linewidth}{!}{
{\small
\begin{tabular}{rccc}\toprule
 Dataset & DIVA Extend & Uniform  & Improvement \\
\midrule
\midrule
Aircrafts &	58.00 & 60.01 &+2.01 \\
Cub-200 & 39.42 &  42.29 & +2.87 \\ 
MIT Indoor-67  &	32.54 &          33.73 & +1.19 \\
Oxford Flowers &20.56 &          23.29 & +2.73 \\
Stanford Cars &	60.37 & 62.45 &         +2.09  \\
Caltech-256	& 21.97 &  24.55 &         +2.59\\
\bottomrule
\end{tabular}
}
}

\caption{\label{tab:diva-extend}Results of using DIVA Extend to select the best samples to extend several fine-grain classification datasets. We train a linear classifier on top of a ResNet-34 pretrained on ImageNet, and compare the test performance when extending the target training dataset with 50\% of the pool samples selected either uniformly at random or via DIVA Extend.
\vspace{-0.5em}
}

\end{table}

\begin{figure}[H]
    \centering
    \includegraphics[width=0.78\linewidth]{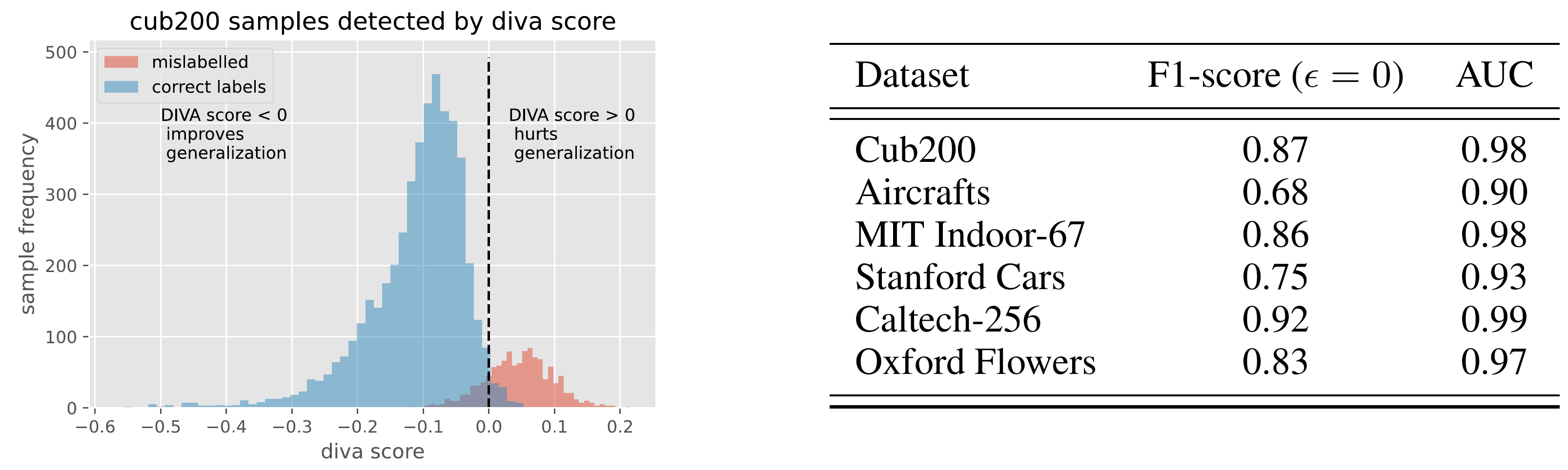}
    \caption{\textbf{(Left)} Distribution of LOO DIVA gradients for correctly labelled and mislabelled samples in CUB-200 dataset (20\% of the samples are mislabeled by replacing their label uniformly at random).
     \textbf{(Right) DIVA for outlier rejection.} We use DIVA on a ResNet-34 network linearization and detect mislabelled samples (outliers) in a dataset present with 20\% label noise. Selection is based on $\nabla_\alpha (L_{\val}(\wb_{\alpha}))_i > \epsilon$.}
     \label{fig:mislabeled-detection}
\end{figure}
  
\paragraph{Detrimental sample detection.}  To test the ability of DIVA to detect training samples that are detrimental for generalization, we artificially introduce wrong labels in the dataset. In \Cref{sec:diva-applications} we claimed that detrimental examples can be detected by looking at the samples for which the derivative $\nabla_\alpha L_\loo(\alpha)_i$ is positive. To verify this, in \Cref{fig:mislabeled-detection} 
we plot the histogram of the derivatives for correct and mislabeled examples. We observe that most mislabeled examples have positive derivative. In particular, we can directly classify an example as mislabeled if the derivative is positive. In \Cref{fig:mislabeled-detection} 
we report the F1 score and AUC obtained in a mislabeled sample detection task using the DIVA gradients.

\begin{figure}[H]

    \vspace{-.5em}

    \centering
    \includegraphics[width=0.9\linewidth]{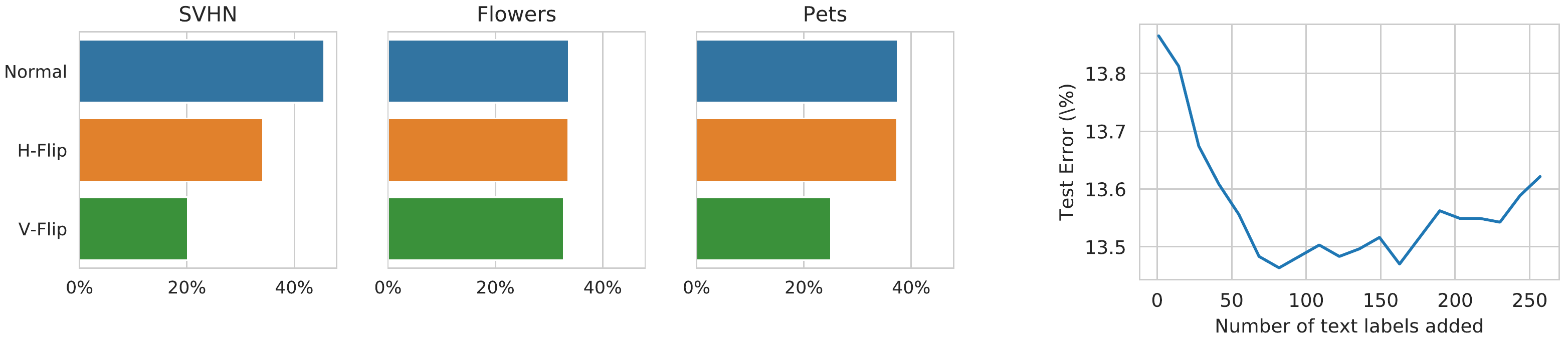}
    \vspace*{-.31em}
    \caption{\textbf{(Left) DIVA detects suitable data augmentation for each task.} We optimize the weight with which each data augmentation transformation is applied. DIVA optimizes the transformation to apply for the particular task.
    \textbf{(Right) Use of DIVA extend on a multi-modal task.} Selecting only the most beneficial text embeddings to use in a multi-modal classification task (as scored by DIVA) outperforms blindly using all available text embeddings.}
    \label{fig:augmentation}
    \vspace*{-.5em}
\end{figure}

\paragraph{Multi-modal learning.} Recent multi-modal models such as CLIP \citep{radford2021learning} can embed both text and images in the same vector spaces. This allows to boost the performance on few-shot image classification tasks by also adding to the training set textual descriptions of the classes, such as the label name.
However, training on label names may also hurt the performance, for example if the label name is not known by the CLIP model. To test this, we create a few-shot task by selecting 20 images per class from the Caltech-256 dataset. We then use DIVA Extend to select an increasing number of labels to add to the training set. In \Cref{fig:augmentation} 
(right), we show that DIVA can select the beneficial label embeddings to add in order to improve the few-shot test performance. However, when forced to add all labels, including detrimental ones, the test error starts to increase again.

\paragraph{Data augmentation.} To further test the versatility of DIVA, we qualitatively evaluate DIVA Reweight on the task of tuning the probabilities with which we apply a given data augmentation procedure. Let $t_1,\ldots,t_K$ be a set of data augmentation transformations. Let $\cD^{t_k}$ be the result of applying the data augmentation $t_k$ to $\cD$. We can create an augmented dataset 
$\cD^\text{aug}=\cD \cup \cD^{t_0} \cup \ldots \cup \cD^{t_K}$,
by merging all transformed datasets. We then apply DIVA Reweight on $\cD^\text{aug}$ to optimize the weight $\alpha$ of the samples. Based on the updated importance weights we estimate the optimal probability with which to apply the transformation $t_k$ as
$p_k = (\sum_{i \in \cD^{t_k}} \alpha_i)/(\sum_i \alpha_i)$. In particular we select common data augmentation procedures, horizontal flip and vertical flip, and tune their weights on the Street View House Number,  Oxford Flowers and the Oxford Pets classification tasks. We observe that DIVA assigns different weights to each transformation depending on the task (\Cref{fig:augmentation}): on the number classification task, DIVA penalizes both vertical and horizontal flips, which may confuse different classes (such 2 and 5, 6 and 9). On an animal classification task (Oxford Pets) DIVA does not penalize horizontal flips, but penalizes vertical flips since they are out of distributions. Finally, on Flowers classification, DIVA gives equal probability to all transformations (most flower pictures are frontal hence rotations and flips are valid).
\vspace*{-\baselineskip}

\section{Discussion}
\label{sec:discussion}

In this work we present a gradient-based method to optimize a dataset. In particular we focus on sample reweighting, extending datasets, and removing outliers from noisy datasets. We note that by developing the notion of a dataset derivative we are capable of improving dataset quality in multiple disparate problems in machine learning. The dataset derivative we present is given in closed-form and enables general reweighting operations on datasets based on desired differentiable validation losses. In cases where a set-aside validation loss is not available we show the use of the leave-one-out framework enables computing and optimizing a dataset ``for free'' and derive the first closed-form dataset derivative based on the LOO framework.

\pagebreak

\bibliography{main}
\pagebreak

\appendix

\section*{Appendix}
We structure the appendix as follows: We present additional experiments in Section \ref{sec:addl_exps} and we describe the details of the experiments in Section \ref{sec:exp_detail}. In Section \ref{sec:proofs} we provide proofs for the propositions in the paper and additional discussion on the methods. 

\section{Additional Experiments}\label{sec:addl_exps}

\paragraph{Validation overfitting.}
When updating the dataset using the dataset derivative there is a risk of overfitting to the validation set after repeated applications of the derivative. Namely the validation loss is initially an unbiased estimate of the test loss yet after using it to update the dataset repeatedly it eventually will start overfitting. In our settings, we notice that when using a small number of gradient updates ($<5$) and with step sizes $\eta \sim 0.1$ we are able to avoid overfitting and improve the \emph{test} error when optimizing the dataset derivative based on the validation loss.
In this experiment we present the final test and validation classification errors of optimized datasets. As we optimize with respect to the validation loss, it is indeed clear that the validation loss decreases dramatically yet more importantly are the effect of the test accuracy.
\begin{figure}[H]
    \centering
    \includegraphics[width=0.4\linewidth]{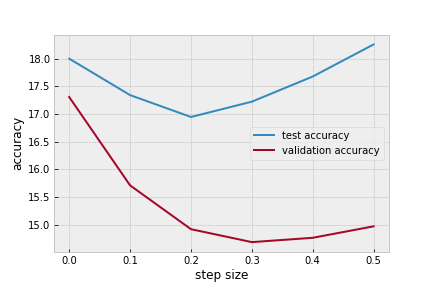}
    \caption{The test error decreases as the dataset is optimized with respect to the validation set until eventually overfitting commences. The validation set error decreases more significantly as the dataset is optimized directly on the validation set, yet for very large step-sizes the first order optimization becomes inaccurate. The plot uses the Caltech-256 dataset to illustrate the overfitting}
    \label{fig:val_overfitting}
\end{figure}

\paragraph{DIVA Extend plots.}
In \Cref{fig:extend_increment} we report the results on all the remaining datasets following the set-up of Figure \ref{fig:extend_increment}. 
 
\begin{figure}[H]
    \centering
     \includegraphics[width=0.32\linewidth]{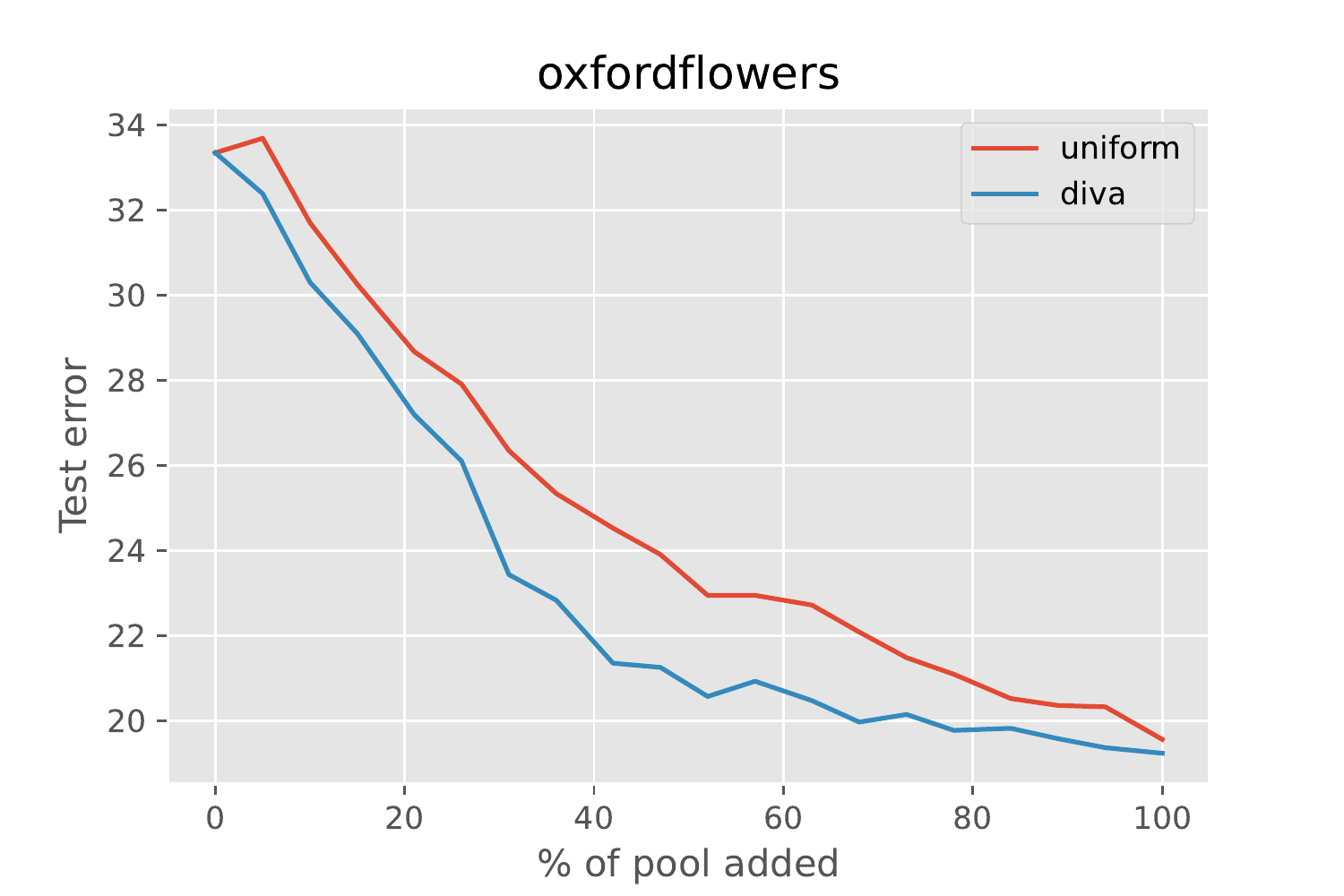} \includegraphics[width=0.32\linewidth]{figures/caltech256_resnet34_19.pdf}
     \includegraphics[width=0.32\linewidth]{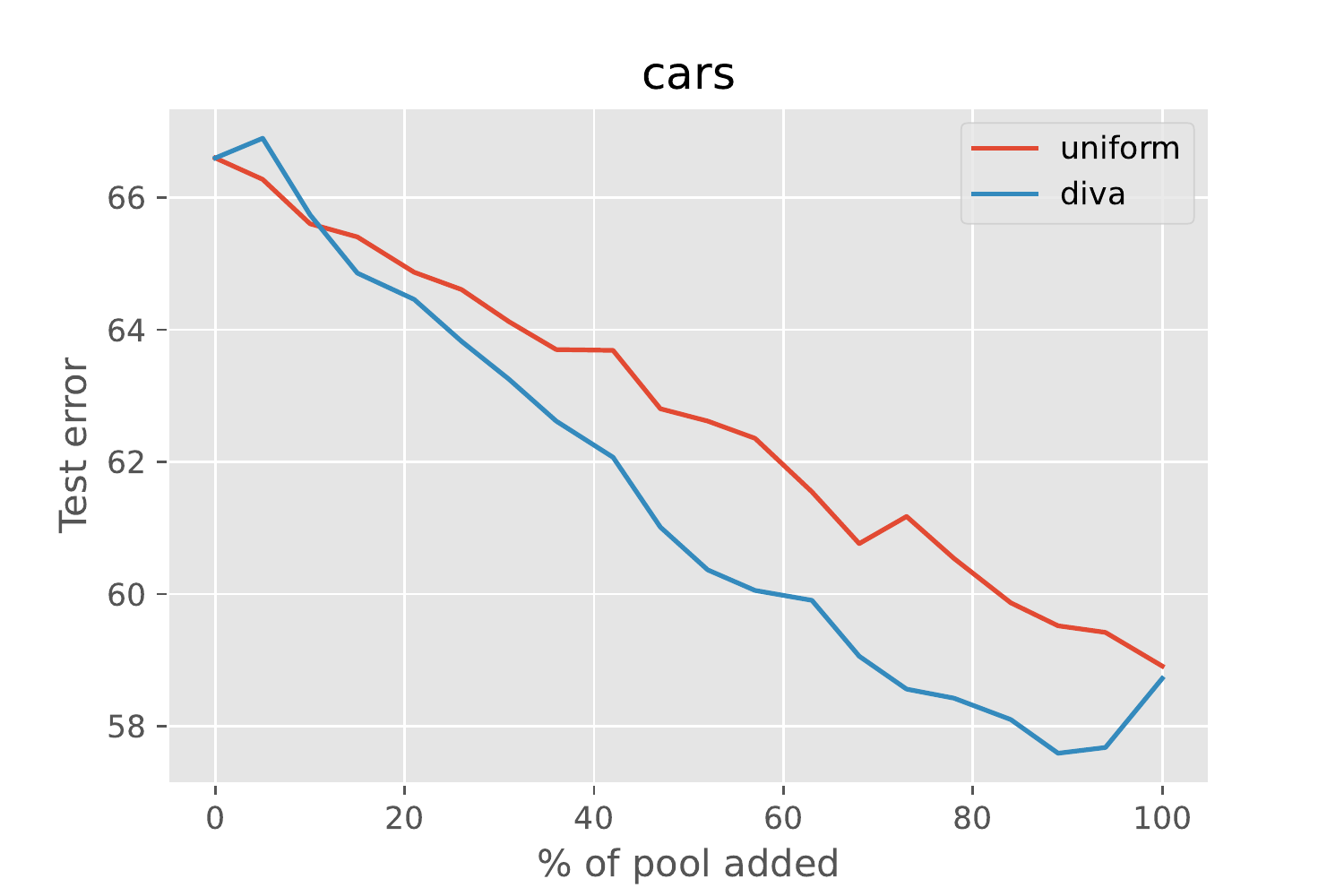}\\
         \centering
     \includegraphics[width=0.32\linewidth]{figures/aircrafts_resnet34_19.pdf} \includegraphics[width=0.32\linewidth]{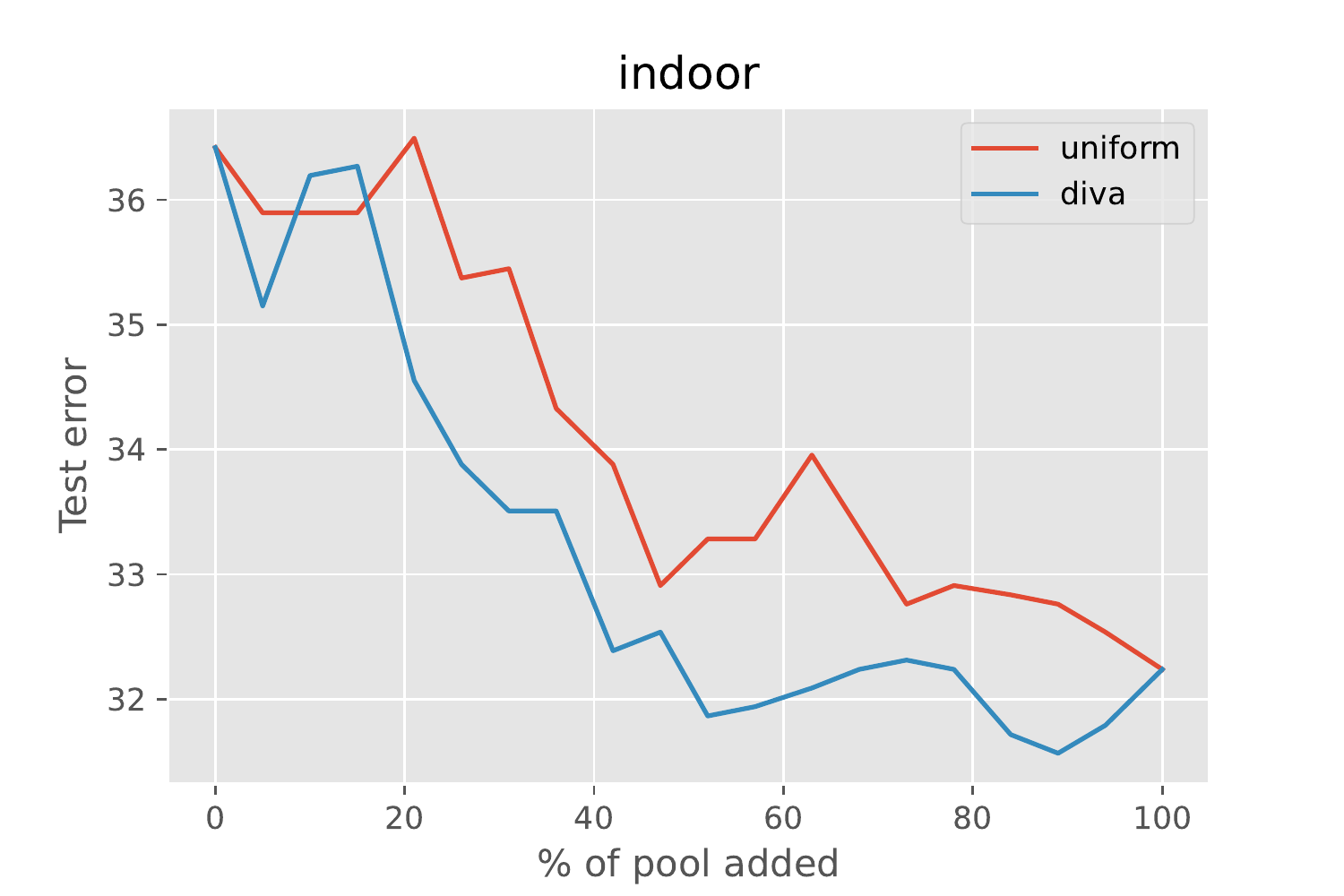}
     \includegraphics[width=0.32\linewidth]{figures/cub200_resnet34_19.pdf}
     \caption{Same plots as \Cref{fig:extend_increment} on the other fine-grained datasets.}
\end{figure}

\section{Experimental details}
\label{sec:exp_detail}

\paragraph{Dataset details} For our experiments we utilize several fine-grain classification datasets from the computer vision community that are standard for fine-tuning image classification tasks (CUB-200 \cite{WelinderEtal2010}, FGVC-Aircraft, \cite{maji13fine-grained}, Stanford Cars \cite{KrauseStarkDengFei-Fei_3DRR2013}, Caltech-256 \cite{griffin2007caltech}, Oxford Flowers 102 \cite{nilsback2008automated}, MIT-67 Indoor \cite{quattoni2009recognizing}, Street View House Number \cite{netzer2011reading}, and the Oxford Pets \cite{parkhi2012cats}).
Some of the datasets do not follow a default train-test split and we use the following splits commonly used in the literature for the datasets,
\begin{itemize}
    \item Oxford Flowers 102: We use the original 1020 images in the training split without incorporating the default validation set.
    \item Caltech-256: We split the dataset into a training set with 60 images from each class for training, and use remaining data for testing.
\end{itemize}
\paragraph{Pre-training setup} For the pre-trained networks we use for fine-tuning, we use the pre-trained configurations available on PyTorch's torchvision package. In particular the reference networks are pre-trained using the ImageNet \cite{deng2009imagenet} dataset. The images embedded by the network are pre-processed via standard resizing and center cropping (256 resize, followed by a 224 cropping).

\paragraph{Regularization parameter $\lambda$}:
To get the best unweighted dataset baseline to compare with the optimized dataset, for each of the un-optimized original datasets, we first search for optimal $\lambda$ values in $\lambda \in \{ 2^n ~\text{for} ~n\in \{-20,-19, \dots 4\} \}$ to measure the classifier's performance. After selecting optimal $\lambda$ we proceed with dataset optimization with the optimal $\lambda$ values. Note that DIVA does not require $\lambda$ to be optimal and improvements are even more significant for un-optimized $\lambda$.

\paragraph{Dataset Derivative Computation}
In Section \ref{sec:proofs} we derive the closed-form dataset derivatives used for DIVA. We computed the closed-form solution analytically and we verified our results using automatic differentiation tools on large number of conditions including synthetic and real data. This has been another method to verify the correctness of the derivative formulas. 

\paragraph{DIVA experimental details}\par
\paragraph{DIVA Extend} For Table \ref{tab:diva-extend} and Figure \ref{fig:extend_increment} we first split the original training set into 50\% training subset and 50\% pool subset that will be used to selectively extend the training subset with DIVA or other extension approaches. We run DIVA Extend LOO and uniform sampling to add pool samples to the training set. In both settings we incrementally extend the training subset from the pool (For DIVA, by selecting samples with top DIVA score) in each step we extend an equal number of samples (($\#$ pool samples) // ($\#$ number of steps)).
In the figure we present the test error as a function of training set size and compare DIVA sample selection with selecting the same number of samples at uniform from the pool set. In Table \ref{tab:diva-extend}, we present the improvement in test accuracy at the 50\% extend mark of the pool set (e.g. extending 25\% of the original training set) between DIVA and uniform sample selection of the same number of samples. For both experiments we use the ResNet-34 architecture.

\paragraph{DIVA Reweigh}
In Table \ref{tab:diva-reweigh-clean} we use DIVA Reweight LOO with the same parameters for all of the datasets we consider. The DIVA parameters we use are $K=4$ for the number of steps and $\eta=0.15$ for the step-size. As with DIVA Extend, we use ResNet 34 for the architecture for the representation. 

\paragraph{DIVA validation loss}
We find the cross entropy loss to work better as the loss function applied to the validation predictions (both in LOO and regular validation). Further for LOO we find it crucial to apply the validation loss only on mis-classified LOO predictions to improve the test accuracy of the model, this can be interpreted as the cross entropy loss with a ``hard margin hinge'' loss.

\paragraph{Outlier rejection} For the results presented in the side-by-side Table and Figure \ref{fig:mislabeled-detection} we apply 20\% random label noise to each class in the dataset and use DIVA LOO to compute the normalized DIVA gradient (DIVA score) of each sample. The F1 score reports classification by thresholding with $\epsilon=0$ and the AUC is computing by thresholds spanning detection of no samples, to detection of all samples.

\section{Proofs of Propositions}\label{sec:proofs}
We repeat the statements of the proposition in the paper for convenience. 
\begin{proposition*}[Model-Dataset Derivative $\nabla_\alpha \wb_\alpha$] For the ridge regression problem \cref{eq:weighted_l2_ridge} and $\wb_{\alpha}$ defined as in \cref{eq:closed_form_sol}, define 
\begin{equation}
\Cb_\alpha = (\Zb^\top \Db_\alpha \Zb + \lambda \Ib)^{-1}.
\end{equation}
Then the Jacobian of $\wb_{\alpha}$ with respect to $\alpha$ is given by
\begin{equation}
    \nabla_{\alpha}\wb_{\alpha} = \Zb \Cb_\alpha \circ \big((\Ib - \Zb \Cb_\alpha \Zb^\top\Db_\alpha)\Yb\big).
\end{equation}
\end{proposition*}
\pp{\ref{prop:model_dataset}}
We recall the settings of the problem are $\Zb \in \RR^{n \times m}$, $\Yb \in \RR^{n\times k}$, $\wb_{\alpha} \in \RR^{m \times k}$ and the importance weights $\alpha \in \RR^{n}$. Here $n$ is the number of samples, $m$ is the number of parameters for each output of the model, and $k$ is the number of classes (represented via the one-hot convention). \par 
The derivation of $\nabla_{\alpha} \wb_{\alpha}$ we present is computational in nature and without the loss of generality we consider the derivative for a single output class $k=1$ (one-vs-all classification naturally extends). For the single class settings, the relevant dimensions are $\wb_{\alpha} \in \RR^{m}$ and $\nabla_{\alpha} \wb_{\alpha} \in \RR^{m \times n}$ (for numerator layout convention of the derivative). 
To further simplify we consider the derivative entrywise for single index $\alpha_r$,
\begin{equation*}
\pdv{\wba}{\alpha_r} \in \RR^{m \times 1}.
\end{equation*}
The closed-form solution is,
\begin{equation*}
     \wb_\alpha = \Cb_{\alpha}\Zb^\top \Db_\alpha \Yb.
\end{equation*}
using chain rule,
\begin{equation*}
    \pdv{\wba}{\alpha_r} = \pdv{\Cb_{\alpha}}{\alpha_r} \times \Zb^\top \Db_\alpha \Yb + \Cb_{\alpha} \times \pdv{\Zb^\top \Db_\alpha \Yb}{\alpha_r}.
\end{equation*}
We compute the two parts of the derivative in turn: \\
Let $\Kb_{\alpha} = (\Zb^\top \Db_{\alpha} \Zb +  \lambda \Ib)$ so that $\Kb_{\alpha} = \Cb_{\alpha}^{-1}$. Then \begin{equation*}
    \pdv{\Kb_{\alpha}}{\alpha_r} = \pdv{}{\alpha_r}(\Zb^\top \Db_{\alpha} \Zb + \lambda \Ib) = \Zb_{r,:}^\top\Zb_{r,:}
\end{equation*} 
where $\Zb_{r,:} \in \RR^{1 \times m}$ is the $r$th row of $\Zb$.
Next note,
\begin{equation*}
    \Cb_{\alpha}\Kb_{\alpha} = \Ib
\end{equation*}
differentiating both sides,
\begin{equation*}
    \pdv{\Cb_{\alpha}\Kb_{\alpha}}{\alpha_r} = \mathbf{0}.
\end{equation*}
Applying chain rule we get,
\begin{equation*}
     \pdv{\Cb_{\alpha}\Kb_{\alpha}}{\alpha_r}=  \pdv{\Cb_{\alpha}}{\alpha_r}\Kb_{\alpha} +  \Cb_{\alpha}\pdv{\Kb_{\alpha}}{\alpha_r},
\end{equation*}
rearranging, substituting $\pdv{\Kb_{\alpha}}{\alpha_r}$, and multiplying to the right by $\Cb_{\alpha}$
\begin{equation*}
    \pdv{\Cb_{\alpha}}{\alpha_r} = -\Cb_{\alpha} \Zb_{r,:}^\top\Zb_{r,:} \Cb_{\alpha}.
\end{equation*}
Next, by direct computation $\pdv{\Zb^\top \Db_\alpha \Yb}{\alpha_r}$ satisfies
\begin{equation*}
    \pdv{\Zb^\top \Db_\alpha \Yb}{\alpha_r} = y_r \cdot \Zb_{r,:}^\top.
\end{equation*}
Combining the original terms we have
\begin{align*}
     \pdv{\wba}{\alpha_r} & = -\big(\Cb_{\alpha} \Zb_{r,:}^\top\Zb_{r,:} \Cb_{\alpha} \big) \times \Zb^\top \Db_\alpha \Yb + y_r \cdot \Cb_{\alpha} \times  \Zb_{r,:}^\top\\
     & = \Cb_{\alpha}\Zb_{r, :}^\top \big(y_r - \Zb_{r,:}\Cb_{\alpha}\Zb^\top\Db_{\alpha}\Yb  \big).
\end{align*}
Now the full derivative is written as
\begin{equation*}
\boxed{
    \nabla_{\alpha}\wb_{\alpha} = \Zb \Cb_\alpha \circ \big((\Ib - \Zb \Cb_\alpha \Zb^\top\Db_\alpha)\Yb\big)}
\end{equation*}

\begin{proposition*}[Validation Loss 
Dataset Derivative]
Define $\Lb$ as the matrix of the loss function derivative with respect to network training outputs as, 
\begin{equation*}
    \Lb = \bigg[\pdv{\ell}{f}(f_{\wba}(\xb_1), y_1), \cdots \pdv{\ell}{f}(f(\xb_N), y_N)\bigg].
\end{equation*}
Then the dataset derivative of the importance weights with respect to final validation loss is given by
\begin{equation}
    \nabla_{\alpha} L_{\val}(\wb_{\alpha}) = \Zb\Cb_\alpha \Zb^\top \times\big( \Lb^\top \Yb^\top (\Ib - \Db_{\alpha}\Zb \Cb_\alpha \Zb^\top)\big).
\end{equation}
\end{proposition*}
\pp{\ref{prop:val_dataset}}
This follows from the chain rule combined with simplification of broadcasting terms.
We again consider the single output settings with the single coordinate derivative $\pdv{L_{\val}}{\alpha_r}$ which is given as,
\begin{equation*}
    \pdv{L_{\val}}{\alpha_r} = \nabla_\wb L_{\val} \pdv{\wba}{\alpha_r}.
\end{equation*}
With 
\begin{align*}
    &= \sum_{i=1}^{n} \Lb_{:,i}\times \zb_i^\top\\
    &= \Lb^\top \Zb.
\end{align*}
Therefore
\begin{equation*}
   \big( \nabla_{\alpha} L_{\val}(\wb_{\alpha})\big)_{i} = \sum_{j,k}\big(\nabla_\alpha \wb_{\alpha} \big)_{i,j,k} (\Lb^\top \Zb)_{j,k}.
\end{equation*}
Now $\Lb, \Zb$ can be separated into the two terms of $\nabla_{\alpha} \wb_{\alpha}$,
\begin{equation}
\boxed{
    \nabla_{\alpha} L_{\val}(\wb_{\alpha}) = \Zb\Cb_\alpha \Zb^\top \times\big( \Lb^\top \Yb^\top (\Ib - \Db_{\alpha}\Zb \Cb_\alpha \Zb^\top)\big)}
\end{equation}

Next we consider the derivations for the Leave One Out (LOO) framework. In the LOO framework one applies cross-validation to a training set
$\{(\zb_1, y_1), \dots (\zb_n, y_n)\}$ by running n-fold cross validation, where in each fold, the $i$th sample $(\zb_i, y_i)$ is taken out and is used for validation while the optimal classifier is solved for the remaining of the training task,
\begin{equation}\label{eq:appendix_loo}  
  \wb_{{-i}},= \arg\min_\wb \sum_{j \ne i} \ell(f(\zb_j), y_j).
\end{equation} 
Then the LOO prediction at the $i$th index is defined as $(\fb_{\text{LOO}})_i = f_{\wb_{-i}}(\zb_i)$. Below we prove the LOO predictions can be written in closed-form without explicit cross validation calculations.

\begin{proposition}[Closed-form LOO prediction vector]\label{prop:loo}
Define the LOO vector predictions as,
\begin{equation*}
\fb_{\loo} = [f_{\wb_{-1}}(\zb_1),\dots, f_{\wb^{-n}}(\zb_n)]^\top
\end{equation*} 
and define \begin{equation*}
    \Rb = \Zb^\top(\Zb^\top\Zb + \lambda \Ib)^{-1}\Zb
\end{equation*}
then for the learning task \cref{eq:appendix_loo} LOO predictions are given as
\begin{equation}
\fb_{\loo} = \frac{\Rb\yb - \diag(\Rb)\yb}{\Ib - \diag(\Rb)}.
\end{equation}
\end{proposition}
\pp{\ref{prop:loo}}
The proof is reproduced from \cite{rifkin2007notes} for completeness. \\
Without the loss of generality we derive the LOO prediction of $\zb_n$. Namely given, $\{(\zb_1, y_1), \dots (\zb_n, y_n)\}$ we use $\{(\zb_1, y_1), \dots (\zb_{n_1}, y_{n-1})\}$ for training and validate using $\{(\zb_n, y_n\}$. 
Denote $\wb^{-n}$ as be the optimal solution to this training task with regularization parameter $\lambda$ and define the (currently unknown) LOO prediction as 
\begin{equation*}
    q = f_{\wb^{-n}}(\zb_n).
\end{equation*}
We define the modified learning task consisting of $\{(\zb_1, y_1), \dots (\zb_{n-1}, y_{n-1}), (\zb_n, q)\}$ where we added the data point $(\zb_n, q)$. Note that the optimal solution with $\lambda$ regularization to the modified learning task is again $\wb^{-n}$ since $q$ is taken to have a zero residual. therefore the solution to the modified learning problem can be written in closed-form as,
\begin{equation*}
    \wb^{-n} =  (\Zb^\top \Zb + \lambda \Ib)^{-1}\Zb^\top [\yb_{1:n-1}, q]^\top .
\end{equation*}
Using $\wb^{-n}$ we write the equation for the LOOV prediction $q$, 
\begin{equation*}
   q = \langle \zb_n,  \wb^{-i} \rangle = \zb_n^\top(\Zb^\top \Zb + \lambda \Ib)^{-1}\Zb^\top [\yb_{1:n-1}, q]^\top. 
\end{equation*}
Let $\Rb = \Zb(\Zb^\top \Zb + \lambda \Ib)^{-1}\Zb^\top$ then we have 
\begin{equation*}
    q = \zb_n^\top \wb_{-i} = \Rb_{n, :} [\Yb_{1:n-1}, q]^\top 
\end{equation*}
and by adding and subtracting $\Rb_{:,n}$ multiplied by $[0, y_n]^\top$ we get,
\begin{equation*}
  q  = \zb_n^\top \wb_{-i} = \Rb_{:n-1, n} [\Yb_{1:n-1}]^\top  + \Rb_{:, n}[0, q] + \Rb_{:,n}[0, y_n]^\top - \Rb_{:,n}[0, y_n]^\top.
\end{equation*}
Re-arranging we have
\begin{equation*}
  q - \Rb_{nn}q =  \Rb_{:, n}\yb - y_{n}\Rb_{n,n}
\end{equation*}
Solving for $q$ we get,
\begin{equation*}
 q=  \frac{\Rb_{:, n}\yb - y_{n}\Rb_{n,n} }{1 - \Rb_{n,n}}
\end{equation*}
And without the loss of generality the full prediction vector is given as,
\begin{equation}
\boxed{
    \fb_{\text{LOO}} = \frac{\Rb\yb - \diag(\Rb) \yb}{\diag(\Ib -\Rb)}}
\end{equation}

Given the closed-form LOOV expression we may use $\fb_{\loo}$ for the validation loss to compute the dataset derivative on $L_{\val}$ without any additional validation data. 
While this may seem contradictory as we are optimizing the dataset validated via the weighting duality between sample loss weighting and data scaling, we define the leave one out value predictions in the weighted dataset settings and evaluate on the original (unweighted) data points as,
\begin{equation}
    f_{\wb_{\alpha}^{-i}}(\zb_i)
\end{equation}
with,
\begin{equation}\label{eq:appendix_weighted_loo}
  \wb_{\alpha}^{-i} = \arg\min_\wb \sum_{j \ne i} \alpha_j \ell(f(\zb_j), y_j). 
\end{equation}    
Therefore the $\alpha$-weighted LOO term is $f_{\wb_{\alpha}^{-i}}(\zb_i)$ is faithful to the original distribution despite being trained with the weighted loss \cref{eq:appendix_weighted_loo}. We in fact are able to show that $\alpha$-weighted LOO formulation also admits a closed-form solution that satisfies our definition and for DIVA LOO we utilize the derivative of the closed-form to optimize the dataset. 

\begin{proposition*}
Define \begin{equation*}
\Rb_{\alpha} = \Zb^\top \sqrt{\Db_{
\alpha}}(\Zb^\top \Db_{\alpha} \Zb + \lambda \Ib)^{-1}\sqrt{\Db_{\alpha}}\Zb
\end{equation*}
Then the $\alpha$-weighted LOOV predictions defined in \cref{eq:weighted_loo_dataset} admit a closed-form solution:
\begin{equation}
    f_{\wb_{\alpha}^{-i}}(\zb_i) = \Bigg[ \frac{\Rb_{\alpha}\sqrt{\Db_{\alpha}}\Yb - \diag(\Rb_{\alpha})\sqrt{\Db_{\alpha}}\Yb}{\diag(\sqrt{\Db_{\alpha}} - \sqrt{\Db_{\alpha}}\Rb_{\alpha})}\bigg]_i,
\end{equation}
Further the LOO dataset derivative is well-defined and satisfies the following gradient condition,
\begin{equation}
    \diag(\nabla_{\alpha} \fb_{\text{LOO}}) = \mathbf{0}.
\end{equation}
\end{proposition*}
The gradient condition $\diag(\nabla_{\alpha} \fb_{\text{LOO}}) = \mathbf{0}$ implies that the LOO prediction at $\zb_i$ does not depend on $\alpha_i$ and ensures that the closed-form solution is well defined in the $\alpha$-weighted settings and is differentiable.\par

\pp{\ref{prop:loo_derivative}}\\
The proof builds on Proposition \ref{prop:loo} for the weighted settings. 

In general since LOO expression describes the weighting problem, it must be shown that the introduction of the weights do not break the argument of Proposition \ref{prop:loo}. 
Considering the optimization problem in \cref{eq:weighted_loo_dataset}. We also use the duality between the loss weights and data scaling to note that $\wb_{\alpha}^-i$ can be derived by considering the unweighted LOO  \cref{eq:appendix_loo} with the modified dataset, 
\begin{equation}
\{(\sqrt{\alpha_1}\zb_1, \sqrt{\alpha_1}y_1), \dots (\sqrt{\alpha_n}\zb_n, \sqrt{\alpha_n}y_n)\}.
\end{equation}
Indeed in this settings with the newly defined data the derivation in Proposition \ref{prop:loo} of the final prediction vector of the LOO entries holds with the same optimal solution $\wb_{\alpha}$ due to the duality between data scaling and loss weights. Nonetheless the closed-form LOO predictions $\fb_{\text{LOO}}$ are evaluated at the data-points $\sqrt{\alpha_i}\zb_i$. Since the model is linear the final predictions at the original data-points of the weighted training settings are written as $\fb_{\text{LOO}}/\sqrt{\alpha}$. 

Noting the weighted training problem can be expressed as $\Tilde{\Yb}= \sqrt{\Db_{\alpha}}\Yb, ~ \Tilde{\Zb}= \sqrt{\Db_{\alpha}}\Zb$ we use Proposition \ref{prop:loo} to write analogously
\begin{equation}
    \Rb_{\alpha} = \Zb^\top \sqrt{\Db_{
\alpha}}(\Zb^\top \Db_{\alpha} \Zb + \lambda \Ib)^{-1}\sqrt{\Db_{\alpha}}\Zb
\end{equation}
and divide by $\sqrt{\alpha}$ by multiplying the denominator by $\sqrt{\Db_{\alpha}}$,
\begin{equation}
    f_{\wb_{\alpha}^{-i}}(\zb_i) = \Bigg[ \frac{\Rb_{\alpha}\sqrt{\Db_{\alpha}}\Yb - \diag(\Rb_{\alpha})\sqrt{\Db_{\alpha}}\Yb}{\diag(\sqrt{\Db_{\alpha}} - \sqrt{\Db_{\alpha}}\Rb_{\alpha})}\bigg]_i,
\end{equation}
Since the derivation at the $i$th index of the weighted LOO prediction, $(\fb_{\text{LOO}})_i = (\wb_{\alpha}^{-i})^\top\zb_i$ is entirely independent of $\alpha_i$, we have
\begin{equation*}
    \pdv{(f_{\wb_{\alpha}^{-i}}(\zb_i))_i}{\alpha_i} = 0.
\end{equation*}
\end{document}